\documentclass[journal]{IEEEtran}

\usepackage{amsmath,amsfonts,amssymb,amsthm,mathtools,bm,dsfont}
\usepackage{siunitx}
\usepackage{graphicx}
\usepackage[table]{xcolor}
\definecolor{bestbg}{HTML}{CADBB7}
\definecolor{sndbg}{HTML}{E6EEDD}
\newcommand{\best}[1]{\cellcolor{bestbg}\textbf{#1}}
\newcommand{\snd}[1]{\cellcolor{sndbg}\underline{#1}}
\usepackage{booktabs}
\usepackage{multirow}
\usepackage{array}
\usepackage{tabularx}
\usepackage{makecell}

\usepackage{algorithm}
\usepackage{algpseudocode}

\usepackage{xspace}
\usepackage{enumitem}
\usepackage{cite}
\usepackage{url}
\usepackage[hidelinks,
            pdftitle={MAVEN-T: Reinforced Heterogeneous Distillation for Real-Time Multi-Agent Trajectory Prediction},
            pdfsubject={Multi-agent trajectory prediction}]{hyperref}

\newcommand{\RR}{\mathbb{R}}

\newcommand{\LL}{\mathcal{L}}
\newcommand{\OO}{\mathcal{O}}
\newcommand{\EE}{\mathbb{E}}

\begin{document}

\title{MAVEN-T: Reinforced Heterogeneous Distillation for Real-Time Multi-Agent Trajectory Prediction}

\author{Wenchang~Duan,
Zhenguo~Gao,
Jinguo~Xian,
Yi~Shi,
\thanks{This work was supported in part by the National Natural Science Foundation of China under Grant 12001365, the National Key Research and Development Program of China under Grant 2022YFE0125300, the Shanghai Municipal Science and Technology Major Project under Grant 2017SHZDZX01, and the Shanghai Jiao Tong University Medicine-Engineering Fund under Grants YG2026LC14 and YG2025QNA46. Corresponding author: Yi Shi.

Wenchang Duan, Zhenguo Gao, and Jinguo Xian are with the School of Mathematical Sciences, Shanghai Jiao Tong University, Shanghai 200240, China (e-mail: duanwenchang@sjtu.edu.cn; gaozheng@sjtu.edu.cn; jgxian@sjtu.edu.cn).

Yi Shi is with the Bio-X Institutes, Key Laboratory for the Genetics of Developmental and Neuropsychiatric Disorders, Shanghai Jiao Tong University, Shanghai 200030, China, and also with the Shanghai Key Laboratory of Psychotic Disorders, and Brain Science and Technology Research Center, Shanghai Jiao Tong University, Shanghai 200030, China (e-mail: yishi@sjtu.edu.cn).}}

\markboth{IEEE TRANSACTIONS ON NEURAL NETWORKS AND LEARNING SYSTEMS}%
{Duan \MakeLowercase{\textit{et al.}}: MAVEN-T: Reinforced Heterogeneous Distillation for Real-Time Multi-Agent Trajectory Prediction}


\maketitle

\begin{abstract}
Trajectory prediction is a key component of autonomous driving systems because future motions determine collision checking, behavior planning, and control decisions. The task remains difficult when dense interactions, heterogeneous agent behaviors, multimodal futures, and limited on-board computation must be handled simultaneously. Existing graph, attention, and generative predictors improve interaction reasoning or uncertainty modeling, but their high-capacity designs are often too expensive for real-time deployment; lightweight predictors and conventional knowledge distillation reduce inference cost, but they usually transfer teacher outputs through static imitation and therefore do not explicitly correct safety-relevant teacher bias. This paper proposes \textbf{MAVEN-T}, a reinforced heterogeneous distillation framework for real-time multi-agent trajectory prediction. A high-capacity teacher first models directed local interactions with a surround-aware graph encoder, combines linear-time temporal filtering with shifted-window spatial attention, and decodes maneuver-specific futures through a sparse Mixture-of-Experts head. A compact GRU--Squeeze-and-Excitation student with a Low-Rank Adapted policy head is then trained by feature-, attention-, and semantic-level distillation. To connect prediction quality with downstream behavior, the distilled student is further refined by Proximal Policy Optimization rewards for collision avoidance, comfort, and progress, while a complexity-aware curriculum and Elastic Weight Consolidation stabilize stage-wise training. Experiments on NGSIM, HighD, MoCAD, Argoverse~2, and the Waymo Open Motion Dataset evaluate accuracy, efficiency, ablation, generalization, robustness, and closed-loop safety. The student achieves 6.2$\times$ parameter compression, 3.7$\times$ inference acceleration, and 14.6\,ms latency on an NVIDIA Jetson AGX Orin while preserving competitive prediction accuracy. The results show that task-level reinforcement can serve as a corrective refinement for deployable trajectory distillation.
\end{abstract}


\begin{figure}[!t]
    \centering
    \includegraphics[width=0.98\columnwidth]{figures/novelty_comparison.pdf}
    \caption{Novelty comparison between generic policy distillation with PPO and MAVEN-T. Generic policy distillation transfers action-policy behavior and uses environment rewards during policy distillation. MAVEN-T transfers a multimodal trajectory predictor through feature, attention, and semantic alignment, and then applies PPO as a task-level refinement driven by safety, comfort, and progress rewards in traffic scenes.}
    \label{fig:novelty_comparison}
\end{figure}

\section{Introduction}
\IEEEPARstart{T}{rajectory} prediction is the problem of inferring the future motion of traffic agents from their observed states and context. In autonomous driving, the predicted trajectories are consumed by collision checking, behavior planning, and model-predictive control; therefore, the predictor must be accurate, robust, and fast enough to operate within the planning cycle~\cite{huang2022comprehensive,LanQizhen9956165,LanQizhen9930673}. This requirement is difficult to satisfy because driving scenes contain directed interactions, heterogeneous motion patterns, uncertain future maneuvers, and strict on-device latency constraints.

Existing trajectory predictors address these factors from different angles. Recurrent and convolutional models provide compact temporal encoders, but they have limited capacity for dense interaction reasoning~\cite{deo2018cs_lstm,messaoud2019nlslstm,deeptrack2022}. Graph and attention models improve relational modeling by encoding social, scene, or heterogeneous-agent interactions~\cite{yang2022gtppo,lv2024ssagcn,chen2024himrae,li2025bpsgcn,mo2022heat}. Generative approaches further model multimodal futures through latent variables, diffusion, or flow-based density estimation~\cite{yang2022gtppo,liang2024stglow,c2ftp2025}. These methods show that interaction structure and uncertainty modeling are essential, but the resulting models can be costly when the number of agents, modes, or relation types increases.

A separate line of work targets deployability. Knowledge distillation can train a compact student from a stronger predictor by matching outputs or intermediate representations~\cite{chen2021s2tnet,liu2021mmtransformer,maplesskd2025}. Selective state-space models and lightweight recurrent encoders also reduce sequence-modeling cost~\cite{gu2024mamba,trajmamba2025,kdmamba2025}. However, most distillation pipelines still regard the teacher as a fixed target. This assumption leaves three limitations. First, output-only imitation can transfer an over-confident teacher mode in rare maneuvers. Second, a static transfer loss applies similar supervision strength to simple lane keeping and dense cut-in scenes, although their interaction uncertainty differs. Third, teacher and student networks are often scaled versions of the same reasoning path, which weakens the benefit of using a high-capacity teacher to supervise a latency-oriented student.

A closely related class of methods combines policy distillation with PPO. Proximal Policy Distillation (PPD), for example, lets the student policy collect rewards during distillation and improves sample efficiency on general reinforcement-learning benchmarks~\cite{spigler2025ppd}. MAVEN-T differs from this policy-distillation setting in three aspects. First, PPD distills action policies, whereas MAVEN-T distills a trajectory-forecasting teacher whose output is a multimodal distribution over future coordinates and maneuver semantics. Second, PPD integrates PPO into policy distillation, whereas MAVEN-T performs supervised multi-granular trajectory distillation first and uses PPO only as a task-level refinement of the compact predictor in a replay-based driving simulator. Third, PPD does not target directed road-agent interactions, heterogeneous traffic participants, on-device latency, or closed-loop safety metrics, while these constraints determine the architecture and rewards in MAVEN-T.

The distinction is illustrated in Fig.~\ref{fig:novelty_comparison}, where the reward signal in MAVEN-T is used to refine a trajectory predictor after structured teacher--student transfer rather than to distill a generic action policy.

MAVEN-T addresses these limitations by treating real-time trajectory prediction as a reinforced heterogeneous distillation problem. The teacher is designed for representation quality: it uses a surround-aware graph encoder for directed local interaction, a hybrid Mamba--shifted-window block for temporal--spatial encoding, and a sparse Mixture-of-Experts decoder for maneuver-conditioned multimodal prediction. The student is designed for deployment: it uses a GRU--Squeeze-and-Excitation encoder and a Low-Rank Adapted policy head. The two networks are connected by multi-granular distillation at feature, attention, and semantic levels. After supervised transfer, the student is refined with Proximal Policy Optimization (PPO) rewards that encode collision avoidance, comfort, and progress. A complexity-aware curriculum and Elastic Weight Consolidation (EWC) are used to reduce unstable updates when training shifts from simple to dense scenarios.

We evaluate MAVEN-T on NGSIM, HighD, MoCAD, Argoverse~2, and the Waymo Open Motion Dataset. The experiments cover prediction accuracy, computational efficiency, component ablation, cross-dataset generalization, robustness to perception and adversarial perturbations, and closed-loop safety in a replay-based simulator. The distilled student achieves 6.2$\times$ parameter compression and 3.7$\times$ inference speedup over the teacher, with 14.6\,ms latency on an NVIDIA Jetson AGX Orin, while remaining close to the teacher in prediction accuracy.

The main contributions are summarized as follows.
\begin{enumerate}[label=\arabic*),leftmargin=*,topsep=2pt,itemsep=2pt]
    \item We formulate real-time multi-agent trajectory prediction as a reinforced heterogeneous distillation problem in which a high-capacity interaction teacher and a compact recurrent student are optimized for different roles.
    \item We design a multi-granular transfer objective that aligns motion features, interaction attention, and maneuver semantics, enabling the student to learn structured knowledge rather than only final trajectory coordinates.
    \item We introduce task-level PPO refinement for trajectory forecasting, applying safety, comfort, and progress rewards after multi-granular teacher--student alignment rather than treating PPO as a generic action-policy distillation objective.
    \item We validate the framework on five public datasets and a replay-based closed-loop simulator, reporting prediction accuracy, on-device latency and memory cost, per-component ablation, robustness, and simulated safety. Codes are open sourced in https://github.com/duanwenchang/MAVEN-T
\end{enumerate}

The remainder of this paper is organized as follows. Section~\ref{sec:related} reviews interaction modeling, multimodal uncertainty, efficient distillation, and task-level refinement. Section~\ref{sec:method} presents MAVEN-T. Section~\ref{sec:experiment} reports the experimental evaluation. Section~\ref{sec:discussion} discusses limitations, and Section~\ref{sec:conclusion} concludes the paper.
\section{Related Work}
\label{sec:related}
This section reviews trajectory-prediction studies according to the modeling assumption that is most relevant to MAVEN-T: interaction structure, multimodal uncertainty, and deployment-oriented training.

https://github.com/duanwenchang/MAVEN-T

\subsection{Interaction and Heterogeneous-Agent Modeling}
Early vehicle trajectory predictors often used recurrent encoder--decoder models with maneuver classes~\cite{deo2018cs_lstm,messaoud2019nlslstm,messaoud2021mhalstm}. Later graph and attention models made interaction structure explicit. GTPPO introduces graph attention with obstacle-avoidance experience and a pseudo-oracle latent variable for future behavior~\cite{yang2022gtppo}. SSAGCN models social and scene interactions with soft attention over graph convolution~\cite{lv2024ssagcn}. HIMRAE focuses on heterogeneous multi-agent systems and analyzes accumulated error from spatial and temporal perspectives~\cite{chen2024himrae}. BP-SGCN further shows that behavioral pseudo-labels can bridge pedestrian-only and heterogeneous-agent prediction without relying solely on manual class labels~\cite{li2025bpsgcn}. These studies motivate directed and heterogeneous interaction modeling; MAVEN-T follows this direction but separates the high-capacity interaction teacher from the compact deployment student.

\subsection{Multimodal Uncertainty Modeling}
Future motion is intrinsically multimodal because agents may choose different lanes, accelerations, and interaction responses. GAN-, CVAE-, and diffusion-based predictors generate diverse futures by sampling latent variables or denoising trajectories~\cite{gupta2018sgan,salzmann2020trajectron,c2ftp2025}. STGlow uses a flow-based generative framework to model the data distribution through exact likelihood and dual graphormer interaction encoding~\cite{liang2024stglow}. Intention-aware predictors such as I2T and IDM decouple maneuver intention from coordinate regression~\cite{zhou2024i2t,idm2025}. These methods improve multimodal quality, but their sampling or high-capacity encoders can conflict with real-time deployment. MAVEN-T keeps a multimodal teacher while distilling its feature, attention, and semantic knowledge into a smaller student.

\subsection{Efficient Distillation and Task-Level Refinement}
Deployment-oriented prediction requires balancing accuracy with latency, memory, and energy. DeepTrack reduces highway-prediction cost through lightweight temporal and depthwise convolution~\cite{deeptrack2022}. Selective state-space models such as Mamba provide linear-time sequence modeling, and recent trajectory variants combine state-space blocks with attention or knowledge transfer~\cite{gu2024mamba,trajmamba2025,kdmamba2025}. Knowledge distillation transfers predictive structure from a teacher to a student through output matching, feature alignment, or map-prior transfer~\cite{chen2021s2tnet,liu2021mmtransformer,maplesskd2025}. However, most distillation objectives are static and do not directly optimize safety-relevant behavior. Interactive prediction and planning studies incorporate reward, game-theoretic, or perceived-safety cues~\cite{rain2021,gameformer2023,liao2025minds,liu2024interactive}. MAVEN-T combines these two lines by first distilling a compact student and then refining it with task-level rewards for collision avoidance, comfort, and progress.

\section{Method: MAVEN-T}
\label{sec:method}

MAVEN-T is built around the same sequence used in the prediction pipeline: define the observed scene, infer interaction-aware teacher representations, transfer the teacher knowledge to a compact student, and refine the student with task-level feedback. This organization links each module to one modeling requirement identified above: directed interaction, multimodal uncertainty, deployable inference, and safety-sensitive refinement.

\subsection{Problem Definition}
\label{sec:problem}
Consider a driving scene observed through an ego-centric coordinate frame. At time $t$, the predictor receives an observation history of length $T_h$ comprising the ego-vehicle state $s_t^{\text{ego}}\!\in\!\RR^{d_e}$, surrounding agents $\mathcal{S}_t=\{s_t^i\}_{i=1}^{N}\!\subset\!\RR^{d_s}$ with $N$ time-varying, and a contextual descriptor $c_t\!\in\!\RR^{d_c}$ encoding lane geometry, speed limits, and traffic states. Each agent state vector is $s_t^i=[x_t^i,y_t^i,v_t^i,\theta_t^i,a_t^i,\text{type}^i]^{\!\top}$, where $(x,y)$ is the planar position, $v$ the speed, $\theta$ the heading, $a$ the longitudinal acceleration, and $\text{type}$ encodes vehicle class. Table~\ref{tab:notation} summarizes the principal notation used throughout the paper.

\begin{table}[!t]
\centering
\caption{Notation summary.}
\label{tab:notation}
\renewcommand{\arraystretch}{1.15}
\footnotesize
\begin{tabularx}{\columnwidth}{lX}
\toprule
Symbol & Description \\
\midrule
$T_h,\,T_f$ & Observation / prediction horizon length \\
$N$ & Number of surrounding agents in scene \\
$\mathcal{O}$ & Observation tensor of length $T_h$ \\
$\hat{\mathcal{Y}}$ & Set of $K$ multimodal predicted trajectories \\
$f_{\theta_T},\,f_{\theta_S}$ & Teacher / student network \\
$\mathcal{G}_{\text{GATv2}}$ & Surround-aware graph encoder \\
$\mathcal{E}_T^{\text{Hybrid}}$ & Hybrid Mamba--SWA temporal--spatial block \\
$\mathcal{D}_T^{\text{MoE}}$ & Mixture-of-Experts decoder \\
$\pi_{\theta_S}$ & LoRA-adapted student policy head \\
$\LL_{\text{low/att/sem}}$ & Multi-granular distillation losses \\
$\LL_{\text{PPO}}$ & Clipped Proximal Policy Optimization loss \\
$\LL_{\text{EWC}}$ & Elastic Weight Consolidation penalty \\
$\mathcal{C}(s)$ & Scenario complexity metric \\
$\mathcal{R}(s_t,a_t)$ & Composite reward (safety/comfort/efficiency) \\
$\hat{A}_t$ & Generalized Advantage Estimate at step $t$ \\
\bottomrule
\end{tabularx}
\end{table}

Given the observation window $\mathcal{O}=\{o_{t-T_h+1},\dots,o_t\}$, the model produces $K$ multimodal future trajectories of length $T_f$ for the ego vehicle and surrounding agents, together with probability scores $\{\pi_k\}_{k=1}^K$:
\begin{equation}
    \hat{\mathcal{Y}} = \big\{(\hat{y}^{1:N}_{t+1:t+T_f,k},\,\pi_k)\big\}_{k=1}^K,\qquad \sum_{k=1}^K \pi_k = 1.
\end{equation}
Throughout this paper we adopt $T_h=15$ steps (3\,s history at 5\,Hz) and $T_f=25$ steps (5\,s prediction at 5\,Hz), consistent with the de facto NGSIM/HighD evaluation protocol~\cite{deo2018cs_lstm,liao2025minds}.

\subsection{Overview of MAVEN-T}
\label{sec:overall_arch}
MAVEN-T is organized into three functional stages rather than a list of independent modules. The \emph{representation stage} trains a high-capacity teacher to encode multi-agent interaction; the \emph{deployment stage} defines a compact student for real-time inference; and the \emph{refinement stage} transfers knowledge and injects task-level feedback. Formally,
\begin{align}
    f_{\theta_T}(\mathcal{O}) &= \mathcal{D}_T^{\text{MoE}}\!\big(\mathcal{E}_T^{\text{Hybrid}}\big(\mathcal{G}_{\text{GATv2}}(\mathcal{S}_t,\mathcal{E}_t),\, \mathcal{O}\big)\big),\\
    f_{\theta_S}(\mathcal{O}) &= \pi_{\theta_S}\!\big(\mathcal{E}_S^{\text{GRU--SE}}\big(\mathcal{S}_t,\, \mathcal{O}\big)\big).
\end{align}
$\mathcal{G}_{\text{GATv2}}$ encodes directed local interactions, $\mathcal{E}_T^{\text{Hybrid}}$ combines linear-time temporal filtering with local spatial attention, and $\mathcal{D}_T^{\text{MoE}}$ allocates decoder capacity to maneuver-specific subspaces. The student uses $\mathcal{E}_S^{\text{GRU--SE}}$ and a Low-Rank Adapted policy head $\pi_{\theta_S}$ to reduce parameters and latency. This separation is deliberate: the teacher is optimized for representation quality, whereas the student is optimized for the deployment budget.

\begin{figure*}[!t]
    \centering
    \includegraphics[width=0.94\textwidth]{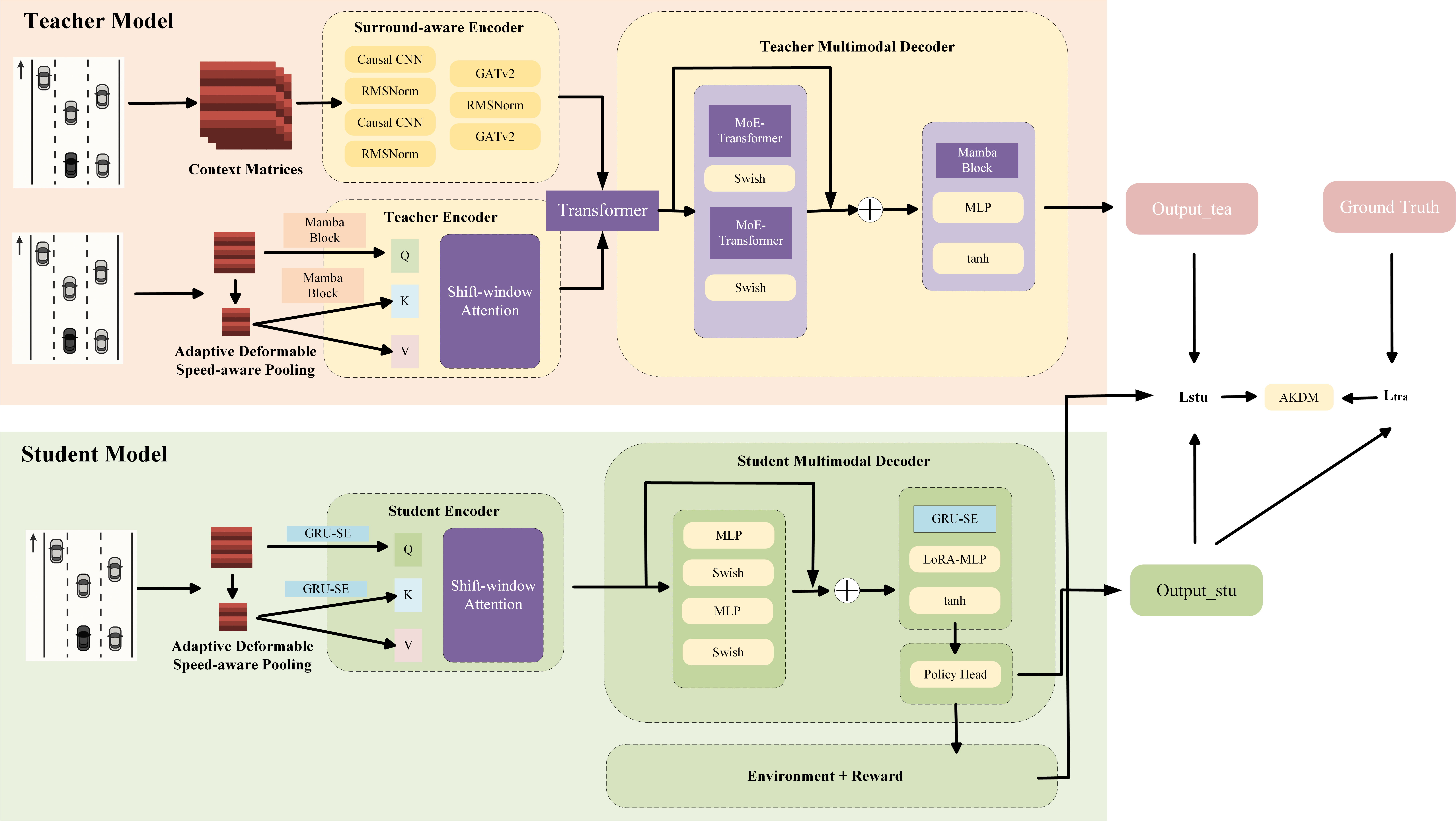}
    \caption{Overview of MAVEN-T. The teacher (top) uses a Surround-Aware GATv2 encoder, a hybrid Mamba--shifted-window block, and a sparse Mixture-of-Experts decoder for accurate interaction reasoning. The student (bottom) uses a GRU--Squeeze-and-Excitation encoder and a Low-Rank Adapted policy head for on-device inference. Multi-granular distillation transfers feature, attention, and semantic knowledge; PPO then refines the student with safety, comfort, and progress rewards under a complexity-aware curriculum.}
    \label{fig:architecture}
\end{figure*}

The remainder of this section follows the data flow in Fig.~\ref{fig:architecture}: interaction graph construction, teacher temporal--spatial encoding, compact student decoding, knowledge transfer, task-level refinement, and complexity analysis.

\subsection{Heterogeneous Interaction Graph Encoder}
\label{sec:surround}
The encoder converts the raw multi-agent state tensor into context-aware embeddings that respect temporal causality and inter-agent topology. In contrast to purely distance-based pooling, the graph keeps directed relation features so that cut-in, following, and lateral-neighbor interactions can receive different attention weights. At each time step $t$, an adjacency matrix $\mathbf{A}_t\!\in\!\RR^{N\times N}$ is initialized from Euclidean proximity:
\begin{equation}
    \mathbf{A}_{t,ij} = \begin{cases}
        \exp\!\big(-d_{ij}^2/(2\sigma^2)\big), & d_{ij} < r_{\text{thresh}},\\
        0, & \text{otherwise},
    \end{cases}
\end{equation}
where $d_{ij}=\|p_t^i - p_t^j\|_2$, the connectivity threshold is $r_{\text{thresh}}=50\,\mathrm{m}$ for highway scenarios, and the kernel bandwidth $\sigma=10\,\mathrm{m}$ controls the spatial decay (so that an interacting neighbor at $30\,\mathrm{m}$ retains roughly $1\%$ of the weight at $0\,\mathrm{m}$). A causal one-dimensional convolution first extracts local temporal patterns:
\begin{equation}
    \mathbf{H}^{(0)} = \mathrm{CausalConv1D}(\mathbf{C}_{1:T_h}),
\end{equation}
where $\mathbf{C}_{1:T_h}\!\in\!\RR^{T_h\times N\times d_c}$ stacks per-step state vectors.

We then apply two stacked Graph Attention Network~v2 (GATv2)~\cite{brody2022gatv2} layers with interleaved RMSNorm~\cite{zhang2019rmsnorm}, replacing the original GAT to alleviate the limited expressivity of static attention scores. The attention coefficient is
\begin{equation}
    \alpha_{ij}^{(l)} = \frac{\exp\!\big(\mathrm{LeakyReLU}(\mathbf{a}^{\!\top}[\mathbf{W}^{(l)}\mathbf{h}_i^{(l)}\!\oplus\!\mathbf{W}^{(l)}\mathbf{h}_j^{(l)}])\big)}{\displaystyle\sum_{k\in\mathcal{N}(i)}\exp\!\big(\mathrm{LeakyReLU}(\mathbf{a}^{\!\top}[\mathbf{W}^{(l)}\mathbf{h}_i^{(l)}\!\oplus\!\mathbf{W}^{(l)}\mathbf{h}_k^{(l)}])\big)},
\end{equation}
with $\oplus$ denoting concatenation. Node features are updated by neighborhood aggregation,
\begin{equation}
    \mathbf{h}_i^{(l+1)} = \sigma\!\Bigg(\sum_{j\in\mathcal{N}(i)}\alpha_{ij}^{(l)}\mathbf{W}^{(l)}\mathbf{h}_j^{(l)}\Bigg),
\end{equation}
and then normalized by RMSNorm,
\begin{equation}
    \mathrm{RMSNorm}(\mathbf{x}) = \frac{\mathbf{x}}{\sqrt{\frac{1}{d}\sum_{i=1}^{d}x_i^2 + \epsilon}}\!\odot\!\mathbf{g},
\end{equation}
with learnable scale $\mathbf{g}$ and $\epsilon=10^{-6}$.

The output is a sequence of context embeddings $\mathbf{F}_{\text{enc}}\!\in\!\RR^{T_h\times d_{\text{model}}}$ shared by the teacher and student paths. Sharing the encoder structurally but not parametrically is critical: it ensures aligned semantic layouts for feature distillation while still letting the two networks learn dimension-specific representations.

\subsection{Teacher Temporal--Spatial Encoder}
\label{sec:hybrid}
\begin{figure}[!t]
    \centering
    \includegraphics[width=0.92\columnwidth]{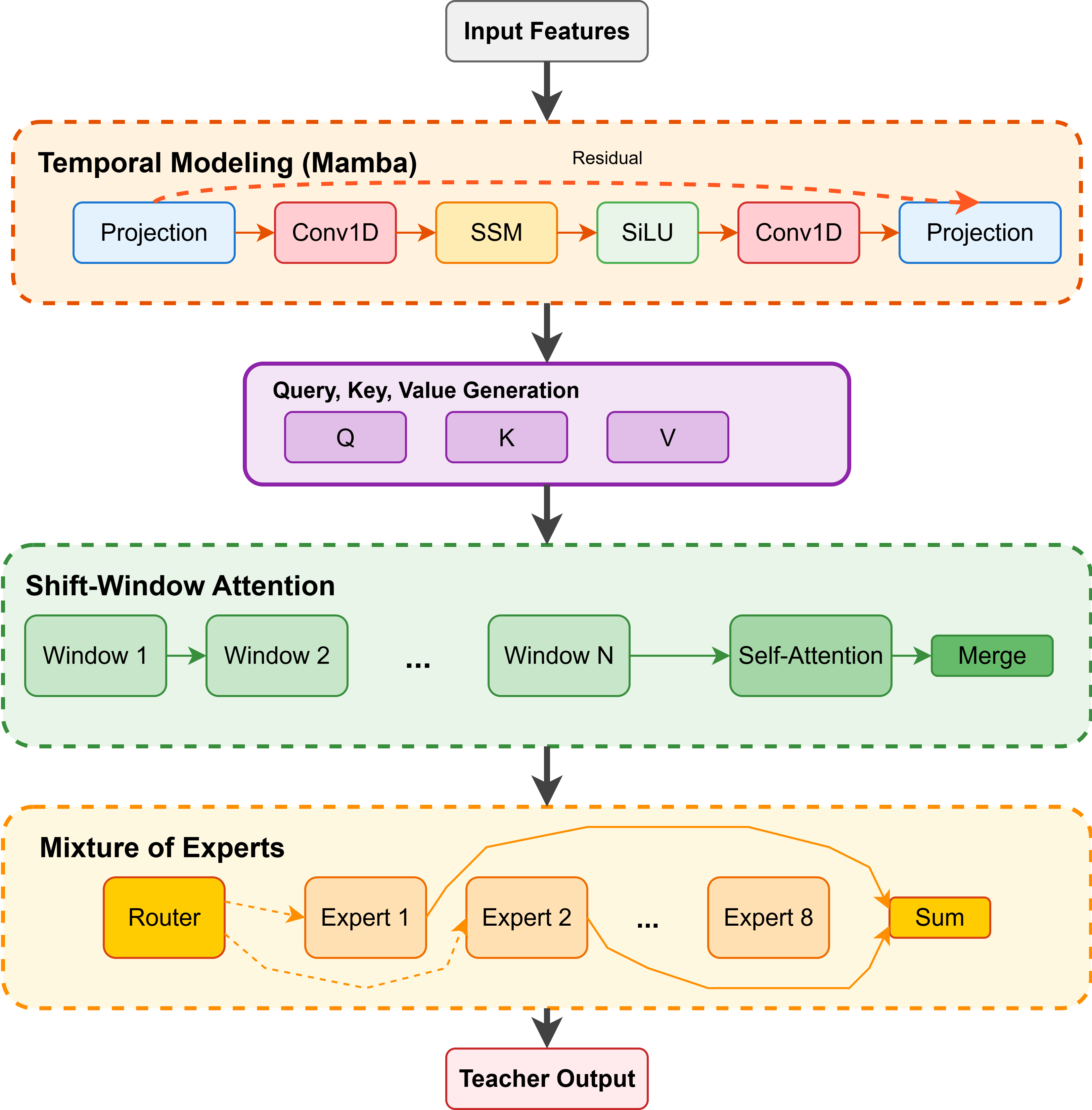}
    \caption{Hybrid Mamba--Shifted-Window Attention block in the teacher network. Linear-time selective state-space modeling captures long-horizon temporal dependencies while windowed self-attention reasons about local spatial interactions; the two streams are fused by a residual gating mechanism.}
    \label{fig:hybrid_attention}
\end{figure}

Highway driving combines two distinct types of dependency. Temporally, the prediction at 5\,s depends on a long history of speed and acceleration in a roughly Markovian manner. Spatially, the relevant interactions are largely local: the most informative neighbors at any moment are within a 30\,m radius. The teacher exploits this dichotomy with a hybrid block (Fig.~\ref{fig:hybrid_attention}).

\subsubsection{Selective state-space temporal modeling}
A Mamba block~\cite{gu2024mamba} replaces the conventional temporal self-attention. For each agent token, the input sequence $\{\mathbf{x}_t\}_{t=1}^{T_h}$ is processed by a continuous-time selective SSM with input-dependent state matrices $(\mathbf{A},\mathbf{B},\mathbf{C})$ and a learnable, input-dependent step size $\Delta\!=\!\Delta(\mathbf{x}_t)$. The zero-order-hold discretization yields
\begin{align}
    \overline{\mathbf{A}} &= \exp(\Delta\,\mathbf{A}),\qquad
    \overline{\mathbf{B}} = (\Delta\,\mathbf{A})^{-1}\!\big(\exp(\Delta\,\mathbf{A})-\mathbf{I}\big)\,\Delta\,\mathbf{B},\\
    \mathbf{h}_t &= \overline{\mathbf{A}}\,\mathbf{h}_{t-1} + \overline{\mathbf{B}}\,\mathbf{x}_t,\qquad
    \mathbf{y}_t = \mathbf{C}\,\mathbf{h}_t + \mathbf{D}\,\mathbf{x}_t.
\end{align}
Here $\Delta\,\mathbf{A}$ and $\Delta\,\mathbf{B}$ denote the products of the scalar step $\Delta$ with the state matrices, not subscripts. This block has $\OO(T_h\,d_{\text{model}})$ complexity, in contrast to the $\OO(T_h^2 d_{\text{model}})$ cost of self-attention, and remains effective for long histories.

\subsubsection{Shifted-window spatial attention}
The Mamba output is reshaped into spatial windows of size $M\!\times\!M$ (with $M\!=\!4$), and a Swin-style~\cite{liu2021swin} self-attention is applied within each window:
\begin{equation}
    \mathrm{Attn}(\mathbf{Q},\mathbf{K},\mathbf{V}) = \mathrm{softmax}\!\left(\frac{\mathbf{Q}\mathbf{K}^{\!\top}}{\sqrt{d_k}} + \mathbf{B}_{\text{pos}}\right)\!\mathbf{V},
\end{equation}
with $\mathbf{B}_{\text{pos}}$ a learned relative-position bias. A shift by $M/2$ in the next layer connects neighboring windows, ensuring information flow across the full agent set while keeping per-step attention cost at $\OO(NM^2)$.

\subsubsection{Residual gated fusion}
The two streams are fused via a gating coefficient $\mathbf{g}=\sigma(\mathbf{W}_g[\mathbf{y}_{\text{Mamba}}\!\oplus\!\mathbf{y}_{\text{Swin}}])$:
\begin{equation}
    \mathbf{z} = \mathbf{g}\odot\mathbf{y}_{\text{Mamba}} + (1-\mathbf{g})\odot\mathbf{y}_{\text{Swin}}.
\end{equation}
Empirically, this gating allows the network to emphasize Mamba's smoother long-range cues in lane-keeping scenes and Swin's sharper local interactions in dense merging scenes (Section~\ref{sec:exp_ablation}).

\subsection{Teacher Decoder and Compact Student}
\label{sec:student}
\subsubsection{Teacher decoder}
The teacher decoder is a sparse Mixture-of-Experts Transformer with $E\!=\!4$ FFN experts and a top-$2$ gating function. Given a token $\mathbf{x}$,
\begin{align}
    G(\mathbf{x}) &= \mathrm{TopK}\!\big(\mathrm{softmax}(\mathbf{x}\mathbf{W}_g),\,k\!=\!2\big),\\
    \mathrm{MoE}(\mathbf{x}) &= \sum_{i=1}^{E} G(\mathbf{x})_i\,E_i(\mathbf{x}),
\end{align}
where $G(\mathbf{x})$ keeps the two largest entries of the softmax over experts and zeros the rest.
Load-balancing regularization~\cite{shazeer2017moe} encourages distinct experts to specialize in qualitatively different scenarios (e.g., free-flow vs.\ congestion). The decoder output is decoded into $K\!=\!6$ trajectory hypotheses and their probabilities.

\subsubsection{Student encoder and policy head}
The student replaces the hybrid block with a two-layer GRU augmented by a Squeeze-and-Excitation (SE) attention~\cite{hu2018se}. For an activation tensor $\mathbf{U}\!\in\!\RR^{T_h\times d}$, the SE block first pools $\mathbf{U}$ along the temporal axis into a channel descriptor $\bar{\mathbf{u}}\!\in\!\RR^d$, then computes an excitation vector through two fully-connected layers $\mathbf{W}_1\!\in\!\RR^{(d/r)\times d}$ and $\mathbf{W}_2\!\in\!\RR^{d\times(d/r)}$:
\begin{align}
    \bar{\mathbf{u}} &= \tfrac{1}{T_h}\sum_{t=1}^{T_h}\mathbf{u}_t,\\
    \mathbf{s} &= \sigma\!\big(\mathbf{W}_2\,\delta(\mathbf{W}_1\bar{\mathbf{u}})\big),\qquad \tilde{\mathbf{U}} = \mathbf{s}\odot\mathbf{U},
\end{align}
where $\delta(\cdot)$ denotes ReLU, $\sigma(\cdot)$ the sigmoid, and the reduction ratio is $r\!=\!16$. The output is projected through a Low-Rank Adapted (LoRA)~\cite{hu2022lora} policy head:
\begin{equation}
    \pi_{\theta_S}(\mathbf{h}) = (\mathbf{W}_0 + \mathbf{B}\mathbf{A})\mathbf{h},\qquad \mathrm{rank}(\mathbf{B}\mathbf{A}) = r_{\text{LoRA}} = 8,
\end{equation}
with $\mathbf{W}_0$ frozen at initialization and $\mathbf{A},\mathbf{B}$ trainable. The use of LoRA allows efficient policy refinement during the reinforcement learning phase without inflating the parameter count.

\begin{figure}[!t]
    \centering
    \includegraphics[width=0.85\columnwidth]{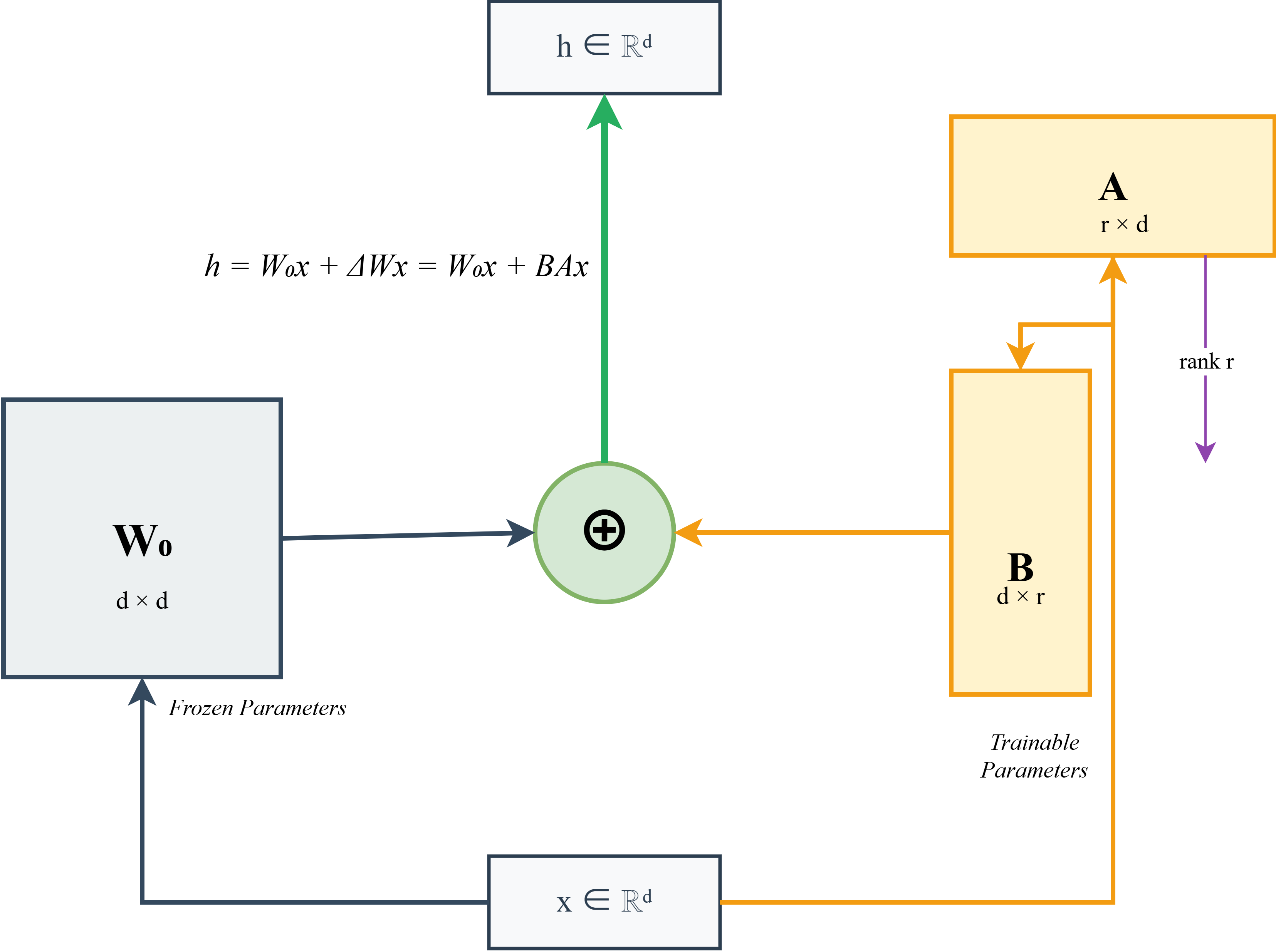}
    \caption{LoRA-parameterized policy head used by the student. Only the low-rank matrices $\mathbf{A}\!\in\!\RR^{r\times d}$ and $\mathbf{B}\!\in\!\RR^{d\times r}$ are updated during reinforcement-augmented distillation, keeping the policy adaptation cost below 1\% of the student's parameter budget.}
    \label{fig:lora_arch}
\end{figure}

The student outputs both a categorical maneuver distribution (lane-keep, left lane change, right lane change) and continuous trajectory parameters, supporting multimodal forecasting with negligible overhead.

\subsection{Multi-Granular Knowledge Transfer}
\label{sec:multigran}
Knowledge transfer between heterogeneous teacher and student is decomposed into three levels.

\paragraph{Low-level feature alignment} Convolutional and SE-pooled features of the student are projected by a $1\!\times\!1$ adapter $\Phi$ to match the teacher's encoder dimensionality and aligned by mean-squared error:
\begin{equation}
    \LL_{\text{low}} = \big\|\mathbf{F}^T_{\text{low}} - \Phi(\mathbf{F}^S_{\text{low}})\big\|_F^2.
\end{equation}

\paragraph{Mid-level attention alignment} The student's GRU hidden-state interaction matrix $\mathbf{A}^S$ is regressed onto the teacher's attention map $\mathbf{A}^T$ (taken from the Swin block) after spatial up-sampling:
\begin{equation}
    \LL_{\text{att}} = \frac{1}{T_h H W}\sum_{t,h,w}\big|\mathbf{A}^T_{t,h,w} - \mathbf{A}^S_{t,h,w}\big|^2.
\end{equation}

\paragraph{High-level semantic alignment} Mode-level latent embeddings $\mathbf{z}^T,\mathbf{z}^S\!\in\!\RR^{d_z}$ are aligned by contrastive InfoNCE,
\begin{equation}
    \LL_{\text{sem}} = -\log\frac{\exp(\mathrm{sim}(\mathbf{z}^T,\mathbf{z}^S)/\tau)}{\sum_{i}\exp(\mathrm{sim}(\mathbf{z}^T,\mathbf{z}^S_i)/\tau)},
\end{equation}
with cosine similarity $\mathrm{sim}(\cdot,\cdot)$ and temperature $\tau\!=\!0.07$, pushing the student to share the teacher's mode-level semantics while distinguishing alternative modes.

\subsection{Task-Level Reinforcement Refinement}
\label{sec:rl}
The student is further fine-tuned in a closed-loop simulator built on a replay of NGSIM, HighD, and MoCAD scenarios augmented by a kinematic bicycle model. The reward is
\begin{equation}
    \mathcal{R}(s_t,a_t) = w_1 R_{\text{safety}} + w_2 R_{\text{comfort}} + w_3 R_{\text{eff}},
\end{equation}
where
\begin{align}
    R_{\text{safety}} &= -\exp\!\left(-\frac{d_{\min}(s_t,a_t)}{d_{\text{safe}}}\right) - \lambda_{\text{viol}}\,\mathds{1}_{\text{violation}},\\
    R_{\text{comfort}} &= -\|a_t-a_{t-1}\|^2 - \mu\,\max(0,|a_t|-a_{\max}),\\
    R_{\text{eff}} &= v_t\cos(\theta_{\text{head}} - \theta_{\text{target}}) - \nu\Delta t,
\end{align}
with $d_{\text{safe}}=\SI{4}{m}$, $a_{\max}=\SI{3}{m/s^2}$, $\lambda_{\text{viol}}\!=\!2.0$. The weights $(w_1,w_2,w_3)\!=\!(0.5,0.3,0.2)$ are determined by Bayesian optimization on a validation split.

We optimize the LoRA-adapted policy with Proximal Policy Optimization (PPO)~\cite{schulman2017ppo}:
\begin{equation}
    \LL_{\text{PPO}}(\theta) = \EE_t\!\Big[\min\!\big(r_t(\theta)\hat{A}_t,\ \mathrm{clip}(r_t(\theta),1-\epsilon,1+\epsilon)\hat{A}_t\big)\Big],
\end{equation}
where $r_t(\theta)=\pi_\theta(a_t|s_t)/\pi_{\theta_{\text{old}}}(a_t|s_t)$, $\hat{A}_t$ is a Generalized Advantage Estimate~\cite{schulman2016gae}, and the clip ratio $\epsilon\!=\!0.2$. Only the LoRA matrices are updated during PPO, which preserves the distilled prior while letting the student correct teacher suboptimalities. When the teacher's expected behavior under a scenario yields negative reward (e.g., overly aggressive cut-ins, undue jerk), PPO updates therefore move the student off the teacher's manifold toward a safer policy rather than re-imitating the teacher. We currently optimize a single-agent policy that conditions on multi-agent observation; related cooperative-RL studies suggest that adaptive context-length truncation with low-frequency filtering~\cite{duan2025aclo} stabilizes multi-agent training under non-stationary partners, which we view as a natural next step for extending MAVEN-T's refinement stage to jointly optimized ego--neighbor policies.

\subsection{Curriculum and Stability Regularization}
\label{sec:curriculum}
\subsubsection{Scenario complexity quantification}
For each training scene $s$ we define a complexity scalar,
\begin{equation}
    \mathcal{C}(s) = 0.3\,\bar{N}_{\text{obj}}(s) + 0.4\,\bar{V}_{\text{rel}}(s) + 0.3\,\mathcal{H}_{\text{traj}}(s),
\end{equation}
where $\bar{N}_{\text{obj}}$ is the mean number of interacting neighbors, $\bar{V}_{\text{rel}}$ the mean relative speed, and $\mathcal{H}_{\text{traj}}$ the trajectory entropy of surrounding agents over the observation window. Weights are tuned by Bayesian optimization on a held-out validation set.

\subsubsection{Progressive curriculum schedule}
Training proceeds in stages indexed by $k$. The current complexity bound advances according to student performance:
\begin{equation}
    \mathcal{C}_{k+1} = \mathcal{C}_k + \Delta\mathcal{C}\cdot \min\!\left(1,\;\frac{\mathrm{Acc}_{\text{stu}} - \mathrm{Acc}_{\text{th}}}{\mathrm{Acc}_{\text{margin}}}\right),
\end{equation}
with $\Delta\mathcal{C}\!=\!0.1$, $\mathrm{Acc}_{\text{th}}\!=\!0.85$, $\mathrm{Acc}_{\text{margin}}\!=\!0.05$.

\subsubsection{Elastic Weight Consolidation}
To prevent catastrophic forgetting between curriculum stages, we add an Elastic Weight Consolidation (EWC)~\cite{kirkpatrick2017ewc} penalty,
\begin{equation}
    \LL_{\text{EWC}} = \sum_i \frac{\lambda}{2}\,\mathbf{F}_i\,(\theta_i - \theta_i^*)^2,
\end{equation}
where $\mathbf{F}_i$ is the diagonal Fisher information at the end of the previous stage and $\theta_i^*$ is the corresponding parameter snapshot.

\subsection{Training Objective and Algorithm}
The full objective for the student is
\begin{equation}
\label{eq:total_loss}
    \LL_{\text{total}} = \alpha_t\LL_{\text{task}} + \xi_t\LL_{\text{low}} + \zeta_t\LL_{\text{att}} + \eta_t\LL_{\text{sem}} - \beta_t\LL_{\text{PPO}} + \psi_t\LL_{\text{EWC}},
\end{equation}
where $\LL_{\text{task}}$ is the standard negative-log-likelihood prediction loss against ground-truth trajectories. Weights $(\alpha_t,\xi_t,\zeta_t,\eta_t,\beta_t,\psi_t)$ follow exponential or saturating schedules:
\begin{equation}
    \xi_t = \xi_0 e^{-\lambda_\xi t},\qquad \beta_t = \beta_0(1-e^{-\lambda_\beta t}),
\end{equation}
mirroring a pedagogical intuition: the student first imitates the teacher heavily and is gradually weaned off as it accumulates RL-driven feedback. The training procedure is summarized in Algorithm~\ref{alg:progressive_distillation}.

\begin{algorithm}[!t]
\caption{Reinforced Multi-Granular Distillation for MAVEN-T}
\label{alg:progressive_distillation}
\begin{algorithmic}[1]
\Require Teacher $T$, Student $S$, dataset $\mathcal{D}$, curriculum depth $K$, simulator $\mathrm{Env}$
\Ensure Distilled student $S^{*}$
\State Pretrain $T$ on $\mathcal{D}$ with NLL until convergence; freeze $T$
\State Initialize complexity $\mathcal{C}_{0}$; stage $k\!\leftarrow\!0$; Fisher $\mathbf{F}\!\leftarrow\!\mathbf{0}$
\For{$k=0$ \textbf{to} $K-1$}
    \State $\mathcal{D}_{k} \leftarrow \mathrm{FilterByComplexity}(\mathcal{D},\mathcal{C}_{k})$
    \Repeat
        \For{minibatch $(\mathcal{O},\mathbf{y})\in\mathcal{D}_{k}$}
            \State $\{\mathbf{F}^T,\mathbf{A}^T,\mathbf{z}^T,\hat{\mathbf{y}}^T\}\!\leftarrow\!T(\mathcal{O})$
            \State $\{\mathbf{F}^S,\mathbf{A}^S,\mathbf{z}^S,\hat{\mathbf{y}}^S\}\!\leftarrow\!S(\mathcal{O})$
            \State Roll out $S$ in $\mathrm{Env}$ for $H$ steps; collect $\{(s,a,r)\}$
            \State Compute $\LL_{\text{total}}$ per Eq.~\eqref{eq:total_loss}
            \State $\theta_S \leftarrow \theta_S - \eta\nabla_{\theta_S}\LL_{\text{total}}$
        \EndFor
        \State $\mathrm{Acc}_{\text{stu}}\!\leftarrow\!\mathrm{Eval}(S,\mathcal{D}_{k}^{\text{val}})$
    \Until{$\mathrm{Acc}_{\text{stu}}\geq \mathrm{Acc}_{\text{th}}$}
    \State $\mathbf{F}\!\leftarrow\!\mathrm{Fisher}(S,\mathcal{D}_{k})$;\; $\theta^*\!\leftarrow\!\theta_S$
    \State $\mathcal{C}_{k+1}\!\leftarrow\!\mathrm{AdvanceCurriculum}(\mathcal{C}_{k},\mathrm{Acc}_{\text{stu}})$
\EndFor
\State \Return $S^{*}\!\leftarrow\!S$
\end{algorithmic}
\end{algorithm}

\subsection{Complexity Analysis}
\label{sec:theory}
The method is designed so that the expensive operations are concentrated in the teacher during offline training. For an observation length $T_h$, agent count $N$, hidden width $d$, and local shifted-window size $w$, the teacher's hybrid block has complexity $\OO(T_hNd)$ for the Mamba temporal scan and $\OO(T_hNwd)$ for local spatial attention, instead of dense $\OO((T_hN)^2d)$ self-attention. The sparse MoE decoder activates only $m$ out of $M$ experts, so its inference cost scales with $m$ rather than $M$. The deployed student removes the MoE and shifted-window attention and uses a GRU--SE encoder whose recurrent cost is $\OO(T_hNd_s^2)$ with a much smaller hidden width $d_s$.

The optimization schedule also follows this separation. Teacher pretraining uses supervised trajectory likelihood; student pretraining uses prediction loss plus multi-granular alignment; PPO refinement is then applied with small clipped updates so that reward optimization does not erase distilled interaction knowledge. EWC penalizes large movement from the previous curriculum stage, and the decaying coefficient $\lambda_{\text{KD}}(t)$ gradually shifts emphasis from teacher imitation to task-level reward.

\section{Experiments}
\label{sec:experiment}
We first describe the datasets, metrics, baselines, and implementation details. We then report prediction accuracy, deployment cost, ablations, cross-dataset transfer, robustness, and closed-loop safety.

\subsection{Experimental Setup}

\subsubsection{Datasets}
We evaluate on five benchmarks widely adopted in vehicle trajectory prediction~\cite{deo2018cs_lstm,mo2022heat,gstcn2022,liao2025minds}:

\textbf{NGSIM}~\cite{ngsim2007}: 10\,Hz Californian-freeway trajectories (US-101 + I-80). 70/10/20 split; 3\,s history $\to$ 5\,s prediction at 5\,Hz~\cite{deo2018cs_lstm}.

\textbf{HighD}~\cite{highd2018}: 25\,Hz drone recordings of 110\,500 trajectories on six bidirectional German highways, split by recording session.

\textbf{MoCAD}~\cite{mocad2024,liao2025minds}: 8.7\,h of Macao Connected Autonomous Driving data (10\,Hz) with complex urban roundabouts.

\textbf{Argoverse~2}~\cite{wilson2023argoverse2}: 250\,000 urban motion-forecasting scenes (10\,Hz) sampled in six U.S.\ cities; 5\,s history $\to$ 6\,s prediction.

\textbf{Waymo Open Motion Dataset (WOMD)}~\cite{ettinger2021womd}: 103\,354 9-second scenes; 1.1\,s history $\to$ 8\,s prediction; we use the standard validation split.

Hyperparameter values used in all experiments are listed in Table~\ref{tab:hyper}.

\begin{table}[!t]
\centering
\caption{Hyperparameters of MAVEN-T.}
\label{tab:hyper}
\renewcommand{\arraystretch}{1.1}
\footnotesize
\begin{tabularx}{\columnwidth}{lXc}
\toprule
Group & Hyperparameter & Value \\
\midrule
Encoder & $d_{\text{model}}$, GATv2 heads, layers & 256, 8, 2 \\
        & RMSNorm $\epsilon$ & $10^{-6}$ \\
        & Interaction radius $r_{\text{thresh}}$ & 50\,m \\
Teacher hybrid & Mamba $d_{\text{state}}$, $d_{\text{conv}}$ & 16, 4 \\
        & Swin window $M$, shift & 4, 2 \\
        & MoE experts $E$, top-$k$ & 4, 2 \\
        & FFN hidden, dropout & 1024, 0.1 \\
Student & GRU layers / hidden & 2 / 256 \\
        & SE reduction ratio $r$ & 16 \\
        & LoRA rank $r_{\text{LoRA}}$ / scaling $\alpha$ & $8$ / $32$ \\
Distill & $\xi_0,\zeta_0,\eta_0$ & 1.0, 0.5, 0.5 \\
        & $\lambda_\xi,\lambda_\zeta,\lambda_\eta$ & 0.02, 0.01, 0.01 \\
        & Contrastive temperature $\tau$ & 0.07 \\
PPO & Clip $\epsilon$, GAE $\lambda$, $\gamma$ & 0.2, 0.95, 0.99 \\
        & Rollout / minibatch & 2048 / 256 \\
        & Reward weights $(w_1,w_2,w_3)$ & (0.5, 0.3, 0.2) \\
Curriculum & $\Delta\mathcal{C}$, $\mathrm{Acc}_{\text{th}}$, stages $K$ & 0.1, 0.85, 5 \\
EWC & $\lambda$ & 400 \\
Optim & Optimizer / LR / WD & AdamW / $3\!\times\!10^{-4}$ / $10^{-4}$ \\
        & Schedule / batch / clip & Cosine / 128 / 1.0 \\
        & Precision / epochs & FP16 / 60+35+15 \\
\bottomrule
\end{tabularx}
\end{table}

\subsubsection{Evaluation metrics}
For deterministic single-modal evaluation we report Root Mean Square Error (RMSE) at 1--5\,s horizons. For multimodal evaluation we report $\mathrm{minADE}_K$, $\mathrm{minFDE}_K$, and Miss Rate $\mathrm{MR}_K$ for $K\!=\!6$, with the additional brier-minFDE~\cite{wilson2023argoverse2} on Argoverse~2 and mAP on WOMD~\cite{ettinger2021womd}. For deployment we additionally report parameter count, GFLOPs, latency, energy consumption, and peak memory. For closed-loop evaluation we report collision rate, off-road rate, jerk RMS, and success rate.

\subsubsection{Baselines}
Eighteen baselines covering five families are evaluated:

\noindent\emph{Recurrent and convolutional}: V-LSTM, S-LSTM~\cite{alahi2016slstm}, CS-LSTM~\cite{deo2018cs_lstm}, MHA-LSTM~\cite{messaoud2021mhalstm}, NLS-LSTM~\cite{messaoud2019nlslstm}, DeepTrack~\cite{deeptrack2022}.

\noindent\emph{Graph- and attention-based}: STDAN~\cite{chen2022stdan}, WSiP~\cite{wsip2023}, GRIP++~\cite{li2019gripplus}, HEAT~\cite{mo2022heat}, HiVT~\cite{zhou2022hivt}, MTR~\cite{shi2024mtr}.

\noindent\emph{Diffusion / intention-aware}: C2F-TP~\cite{c2ftp2025}, I2T~\cite{zhou2024i2t}, NEST~\cite{nest2025}, IDM~\cite{idm2025}.

\noindent\emph{Knowledge-distilled / cognitive}: HLTP~\cite{liao2024hltp}, CITF~\cite{liao2025minds}.

\noindent\emph{State-space}: Trajectory-Mamba~\cite{trajmamba2025}.

For the lightweight student, we additionally compare against MobileNet-Traj, DistilBERT-Traj, and Lightweight-LSTM as compression-oriented baselines.

\subsubsection{Implementation details}
The teacher and student are implemented in PyTorch~2.1 with CUDA~12.1. Training uses AdamW with cosine annealing from $3\!\times\!10^{-4}$ to $1\!\times\!10^{-5}$, batch size 128, gradient clip 1.0, and FP16 mixed precision. The teacher is trained for 60 epochs on a single NVIDIA A100~80\,GB; the student is then trained for 35 distillation epochs followed by 15 PPO epochs. The simulator uses a kinematic bicycle model with a 0.1\,s integration step, replayed in batches of 32 parallel environments. Experiments use five random seeds, and reported confidence intervals are computed from seed-level results. For reproducibility, all methods use the same dataset splits, preprocessing pipeline, checkpoint-selection rule, and latency measurement protocol.

\subsection{Comparison With Existing Methods}

\subsubsection{NGSIM and HighD}
Table~\ref{tab:teacher_rmse} reports RMSE on NGSIM and HighD. MAVEN-T (Teacher) attains the lowest RMSE at every prediction horizon: averaged across 1--5\,s it improves over CITF~\cite{liao2025minds} by 5.6\% on NGSIM and 6.0\% on HighD, and over the diffusion-based C2F-TP~\cite{c2ftp2025} by 5.4\% and 6.7\% respectively. The MAVEN-T \emph{student}---despite having 6.2$\times$ fewer parameters than the teacher---remains within 2.3\% of the teacher's accuracy and outperforms HLTP~\cite{liao2024hltp} (the closest distilled competitor) by 9.4\% on NGSIM and 11.2\% on HighD in average RMSE.

\begin{table*}[!t]
\centering
\caption{RMSE (m) on NGSIM and HighD across 1--5\,s prediction horizons. Best in \textbf{bold}, second best \underline{underlined}.}
\label{tab:teacher_rmse}
\renewcommand{\arraystretch}{1.1}
\footnotesize
\begin{tabular}{lccccc|c||ccccc|c}
\toprule
\multirow{2}{*}{Method} & \multicolumn{6}{c||}{NGSIM} & \multicolumn{6}{c}{HighD} \\
& 1\,s & 2\,s & 3\,s & 4\,s & 5\,s & Avg & 1\,s & 2\,s & 3\,s & 4\,s & 5\,s & Avg \\
\midrule
V-LSTM~\cite{deo2018cs_lstm} & 0.68 & 1.66 & 2.96 & 4.56 & 5.44 & 3.06 & 0.22 & 0.65 & 1.32 & 2.22 & 3.43 & 1.57 \\
S-LSTM~\cite{alahi2016slstm} & 0.59 & 1.29 & 2.13 & 3.21 & 4.55 & 2.35 & 0.21 & 0.65 & 1.31 & 2.16 & 3.29 & 1.52 \\
CS-LSTM~\cite{deo2018cs_lstm} & 0.58 & 1.27 & 2.11 & 3.19 & 4.53 & 2.34 & 0.24 & 0.68 & 1.26 & 2.15 & 3.31 & 1.53 \\
MHA-LSTM~\cite{messaoud2021mhalstm} & 0.41 & 1.01 & 1.74 & 2.67 & 3.83 & 1.93 & 0.19 & 0.55 & 1.10 & 1.84 & 2.78 & 1.29 \\
NLS-LSTM~\cite{messaoud2019nlslstm} & 0.56 & 1.22 & 2.02 & 3.03 & 4.30 & 2.23 & 0.20 & 0.57 & 1.14 & 1.90 & 2.91 & 1.34 \\
DeepTrack~\cite{deeptrack2022} & 0.47 & 1.08 & 1.83 & 2.75 & 3.89 & 2.00 & 0.16 & 0.49 & 1.02 & 1.78 & 2.75 & 1.24 \\
GRIP++~\cite{li2019gripplus} & 0.38 & 0.89 & 1.45 & 2.14 & 2.94 & 1.56 & 0.17 & 0.48 & 0.98 & 1.65 & 2.51 & 1.16 \\
STDAN~\cite{chen2022stdan} & 0.42 & 1.01 & 1.69 & 2.56 & 3.67 & 1.87 & 0.15 & 0.45 & 0.94 & 1.68 & 2.58 & 1.16 \\
WSiP~\cite{wsip2023} & 0.56 & 1.23 & 2.05 & 3.08 & 4.34 & 2.25 & 0.20 & 0.60 & 1.21 & 2.07 & 3.14 & 1.44 \\
HiVT~\cite{zhou2022hivt} & 0.36 & 0.85 & 1.40 & 2.10 & 2.90 & 1.52 & 0.13 & 0.43 & 0.95 & 1.69 & 2.65 & 1.17 \\
HEAT~\cite{mo2022heat} & 0.40 & 0.96 & 1.61 & 2.40 & 3.45 & 1.76 & 0.15 & 0.47 & 1.00 & 1.75 & 2.71 & 1.22 \\
MTR~\cite{shi2024mtr} & 0.33 & 0.84 & 1.42 & 2.09 & 2.92 & 1.52 & 0.12 & 0.42 & 0.93 & 1.65 & 2.58 & 1.14 \\
I2T~\cite{zhou2024i2t} & 0.34 & 0.83 & 1.39 & 2.07 & 2.87 & 1.50 & 0.12 & 0.42 & 0.93 & 1.66 & 2.59 & 1.14 \\
C2F-TP~\cite{c2ftp2025} & 0.32 & 0.92 & 1.62 & 2.44 & 3.45 & 1.75 & 0.11 & 0.41 & 0.92 & 1.64 & 2.60 & 1.14 \\
NEST~\cite{nest2025} & 0.31 & 0.86 & 1.55 & 2.34 & 3.32 & 1.68 & 0.11 & 0.40 & 0.91 & 1.62 & 2.55 & 1.12 \\
IDM~\cite{idm2025} & 0.34 & 0.88 & 1.49 & 2.20 & 3.06 & 1.59 & 0.12 & 0.41 & 0.92 & 1.65 & 2.57 & 1.13 \\
Trajectory-Mamba~\cite{trajmamba2025} & 0.32 & 0.84 & 1.42 & 2.12 & 2.96 & 1.53 & 0.11 & 0.41 & 0.91 & 1.62 & 2.54 & 1.12 \\
HLTP~\cite{liao2024hltp} & 0.36 & 0.92 & 1.58 & 2.36 & 3.34 & 1.71 & 0.13 & 0.45 & 0.97 & 1.71 & 2.66 & 1.18 \\
CITF~\cite{liao2025minds} & \snd{0.31} & \snd{0.83} & \snd{1.41} & \snd{2.13} & \snd{3.10} & \snd{1.56} & \snd{0.10} & \snd{0.38} & \snd{0.87} & \snd{1.58} & \snd{2.49} & \snd{1.08} \\
\textbf{MAVEN-T (Teacher)} & \best{0.28} & \best{0.78} & \best{1.34} & \best{2.02} & \best{2.94} & \best{1.47} & \best{0.09} & \best{0.35} & \best{0.83} & \best{1.52} & \best{2.40} & \best{1.04} \\
\textbf{MAVEN-T (Student)} & 0.30 & 0.80 & 1.37 & 2.06 & 3.00 & 1.50 & 0.10 & 0.37 & 0.85 & 1.55 & 2.44 & 1.06 \\
\bottomrule
\end{tabular}
\end{table*}

\subsubsection{MoCAD}
Table~\ref{tab:mocad} reports performance on MoCAD. The challenging urban roundabouts and shared-space interactions in MoCAD highlight the value of explicit interaction modeling: MAVEN-T (Teacher) reduces RMSE by 8.7\% over CITF and 14.1\% over HLTP, while the student remains within 3.2\% of the teacher.

\begin{table}[!t]
\centering
\caption{RMSE (m) on MoCAD across 1--5\,s.}
\label{tab:mocad}
\renewcommand{\arraystretch}{1.1}
\footnotesize
\begin{tabular}{lcccccc}
\toprule
Method & 1\,s & 2\,s & 3\,s & 4\,s & 5\,s & Avg \\
\midrule
CS-LSTM~\cite{deo2018cs_lstm} & 1.45 & 1.98 & 2.94 & 3.56 & 4.49 & 2.88 \\
MHA-LSTM~\cite{messaoud2021mhalstm} & 1.25 & 1.48 & 2.57 & 3.22 & 4.20 & 2.54 \\
STDAN~\cite{chen2022stdan} & 0.62 & 1.46 & 2.52 & 3.66 & 4.71 & 2.59 \\
WSiP~\cite{wsip2023} & 0.70 & 1.70 & 2.56 & 3.47 & 4.71 & 2.63 \\
HLTP~\cite{liao2024hltp} & 0.55 & 1.39 & 2.28 & 3.18 & 4.19 & 2.32 \\
NEST~\cite{nest2025} & 0.49 & 1.22 & 2.02 & 2.94 & 3.85 & 2.10 \\
CITF~\cite{liao2025minds} & \snd{0.42} & \snd{1.06} & \snd{1.87} & \snd{2.69} & \snd{3.53} & \snd{1.91} \\
\textbf{MAVEN-T (Teacher)} & \best{0.39} & \best{0.97} & \best{1.72} & \best{2.45} & \best{3.20} & \best{1.75} \\
\textbf{MAVEN-T (Student)} & 0.41 & 1.00 & 1.78 & 2.51 & 3.31 & 1.80 \\
\bottomrule
\end{tabular}
\end{table}

\subsubsection{Argoverse~2 and WOMD}
Table~\ref{tab:av2_womd} reports performance on the urban Argoverse~2 (AV2) and WOMD validation splits. On AV2 MAVEN-T reduces brier-minFDE$_6$ by 4.8\% over MTR and 2.1\% over Trajectory-Mamba; on WOMD it lifts mAP from 0.434 (HiVT) to 0.460 (Teacher) / 0.453 (Student). The strong performance under genuinely urban conditions confirms that the hybrid Mamba--Swin encoder generalizes beyond highway traffic.

\begin{table}[!t]
\centering
\caption{Argoverse 2 and Waymo Open Motion Dataset (validation splits).}
\label{tab:av2_womd}
\renewcommand{\arraystretch}{1.1}
\setlength{\tabcolsep}{3pt}
\scriptsize
\begin{tabular}{lcccc}
\toprule
\multirow{2}{*}{Method} & \multicolumn{2}{c}{Argoverse~2} & \multicolumn{2}{c}{WOMD} \\
\cmidrule(lr){2-3}\cmidrule(lr){4-5}
& mFDE$_6\!\downarrow$ & b-mFDE$_6\!\downarrow$ & mAP$\!\uparrow$ & mADE$_6\!\downarrow$ \\
\midrule
HiVT~\cite{zhou2022hivt} & 1.41 & 1.93 & 0.434 & 0.66 \\
Wayformer~\cite{nayakanti2022wayformer} & 1.45 & 1.99 & 0.419 & 0.65 \\
MTR~\cite{shi2024mtr} & 1.29 & 1.82 & 0.448 & 0.60 \\
Trajectory-Mamba~\cite{trajmamba2025} & 1.27 & 1.77 & 0.452 & 0.59 \\
HLTP~\cite{liao2024hltp} & 1.36 & 1.88 & 0.440 & 0.62 \\
CITF~\cite{liao2025minds} & \snd{1.25} & \snd{1.75} & \snd{0.456} & \snd{0.58} \\
\textbf{MAVEN-T (Teacher)} & \best{1.21} & \best{1.73} & \best{0.460} & \best{0.56} \\
\textbf{MAVEN-T (Student)} & 1.24 & 1.76 & 0.453 & 0.58 \\
\bottomrule
\end{tabular}
\end{table}

\subsubsection{Multimodal evaluation}
Table~\ref{tab:multimodal} reports $\mathrm{minADE}_6$, $\mathrm{minFDE}_6$, and $\mathrm{MR}_6$ at 3\,s on NGSIM. MAVEN-T's $K=6$ output recovers the diverse intentions of lane-keeping, left-change, and right-change, while the lower miss rate (3.4\%) confirms the discriminative power of the MoE decoder.

\begin{table}[!t]
\centering
\caption{Multimodal prediction quality on NGSIM at 3\,s ($K=6$). MR threshold 2\,m.}
\label{tab:multimodal}
\renewcommand{\arraystretch}{1.1}
\footnotesize
\begin{tabular}{lccc}
\toprule
Method & $\mathrm{minADE}_6\!\downarrow$ & $\mathrm{minFDE}_6\!\downarrow$ & $\mathrm{MR}_6\!\downarrow$ \\
\midrule
MTR~\cite{shi2024mtr} & 0.61 & 1.18 & 0.061 \\
HiVT~\cite{zhou2022hivt} & 0.66 & 1.27 & 0.067 \\
C2F-TP~\cite{c2ftp2025} & 0.58 & 1.10 & 0.052 \\
HLTP~\cite{liao2024hltp} & 0.63 & 1.21 & 0.063 \\
CITF~\cite{liao2025minds} & 0.55 & 1.04 & 0.045 \\
\textbf{MAVEN-T (Teacher)} & \best{0.51} & \best{0.95} & \best{0.034} \\
\textbf{MAVEN-T (Student)} & \snd{0.53} & \snd{0.99} & \snd{0.038} \\
\bottomrule
\end{tabular}
\end{table}

\subsubsection{Per-maneuver breakdown}
Table~\ref{tab:per_maneuver} disaggregates RMSE by maneuver class on NGSIM. MAVEN-T gains are largest on the difficult left- and right-lane-change classes, with 14.0\% and 12.6\% improvements over HLTP, owing to the MoE decoder's mode specialization.

\begin{table}[!t]
\centering
\caption{Per-maneuver RMSE (m) at 5\,s on NGSIM.}
\label{tab:per_maneuver}
\renewcommand{\arraystretch}{1.1}
\footnotesize
\begin{tabular}{lccc}
\toprule
Method & Lane-Keep & Left-LC & Right-LC \\
\midrule
CS-LSTM~\cite{deo2018cs_lstm} & 3.62 & 5.45 & 5.28 \\
STDAN~\cite{chen2022stdan} & 2.78 & 4.41 & 4.27 \\
WSiP~\cite{wsip2023} & 3.41 & 5.05 & 4.84 \\
HLTP~\cite{liao2024hltp} & 2.43 & 3.85 & 3.71 \\
CITF~\cite{liao2025minds} & 2.24 & 3.52 & 3.40 \\
\textbf{MAVEN-T (Teacher)} & \best{2.07} & \best{3.28} & \best{3.21} \\
\textbf{MAVEN-T (Student)} & \snd{2.12} & \snd{3.31} & \snd{3.24} \\
\bottomrule
\end{tabular}
\end{table}

\subsection{Computational Efficiency and On-Device Deployment}
Table~\ref{tab:efficiency} compares parameter count, FLOPs, server-side latency (NVIDIA A100), and on-device latency on an NVIDIA Jetson AGX Orin under FP16 with batch size 1. MAVEN-T's student attains 1.3\,M parameters and 14.6\,ms latency on the Jetson, comfortably within the 50\,ms budget commonly required by Level-4 planning stacks~\cite{liao2024hltp,liu2024interactive}. The teacher itself is 32\% smaller than C2F-TP yet 5.4\% more accurate, owing largely to the linear-time Mamba block and the sparse MoE decoder.

\begin{table}[!t]
\centering
\caption{Computational efficiency at $T_h=15$, $T_f=25$, $N=20$ agents. Latency reported on NVIDIA A100 (FP16) and Jetson AGX Orin (FP16, batch 1).}
\label{tab:efficiency}
\renewcommand{\arraystretch}{1.1}
\setlength{\tabcolsep}{3pt}
\scriptsize
\begin{tabular}{lcccc}
\toprule
Method & Params (M) & FLOPs (G) & A100 (ms) & Orin (ms) \\
\midrule
\multicolumn{5}{l}{\textbf{Heavy / Teacher class}} \\
STDAN~\cite{chen2022stdan} & 8.5 & 12.3 & 14.2 & 45.2 \\
WSiP~\cite{wsip2023} & 6.8 & 9.8 & 12.1 & 38.7 \\
HiVT~\cite{zhou2022hivt} & 9.4 & 13.6 & 15.4 & 49.6 \\
C2F-TP~\cite{c2ftp2025} & 12.1 & 15.7 & 18.7 & 52.6 \\
MTR~\cite{shi2024mtr} & 65.8 & 38.6 & 32.4 & 89.2 \\
CITF~\cite{liao2025minds} & \snd{11.5} & 14.2 & 16.9 & 48.5 \\
\textbf{MAVEN-T (Teacher)} & \best{8.2} & \best{10.4} & \best{11.8} & \best{36.4} \\
\midrule
\multicolumn{5}{l}{\textbf{Student / Light class}} \\
MobileNet-Traj & 1.8 & 2.1 & 5.7 & 12.5 \\
DistilBERT-Traj & 2.3 & 3.2 & 6.4 & 15.8 \\
Lightweight-LSTM & 1.5 & 1.8 & 4.5 & 11.2 \\
HLTP (student)~\cite{liao2024hltp} & \snd{2.4} & \snd{2.7} & \snd{6.0} & \snd{14.9} \\
\textbf{MAVEN-T (Student)} & \best{1.3} & \best{2.3} & \best{5.1} & \best{14.6} \\
\midrule
\textbf{Teacher$\to$Student} & \textbf{6.2$\times$} & \textbf{4.5$\times$} & \textbf{2.3$\times$} & \textbf{3.7$\times$} \\
\bottomrule
\end{tabular}
\end{table}

\subsubsection{Per-module inference profiling}
Table~\ref{tab:module_profile} breaks down the teacher and student inference latency by module on the Jetson AGX Orin. The hybrid Mamba--Swin block accounts for 41\% of the teacher's latency, while the MoE decoder contributes 27\% despite activating only 2 of 4 experts per token. In the student, the GRU--SE encoder dominates (61\%) while the LoRA-adapted policy head adds only 1.2\,ms.

\begin{table}[!t]
\centering
\caption{Per-module Jetson AGX Orin latency (ms, FP16, batch 1).}
\label{tab:module_profile}
\renewcommand{\arraystretch}{1.1}
\footnotesize
\begin{tabular}{lcc}
\toprule
Module & Teacher (ms) & Student (ms) \\
\midrule
Surround-aware GATv2 encoder & 6.8 & 3.4 \\
Hybrid Mamba--SWA block & 14.9 & --- \\
GRU--SE encoder & --- & 8.9 \\
MoE decoder / LoRA head & 9.8 & 1.2 \\
Output projection \& head & 4.9 & 1.1 \\
\midrule
Total & 36.4 & 14.6 \\
\bottomrule
\end{tabular}
\end{table}

\subsubsection{Inference time vs. number of agents}
The student's inference time scales near-linearly with the number of surrounding agents $N$, from 11.2\,ms at $N\!=\!5$ to 18.4\,ms at $N\!=\!30$ on the Jetson AGX Orin. By comparison, C2F-TP grows from 48.7\,ms to 84.1\,ms in the same range, owing to its quadratic attention complexity. The detailed scaling curves are reported in Table~\ref{tab:agents_scaling}.

\begin{table}[!t]
\centering
\caption{Inference latency (ms) vs.\ number of agents $N$ on Jetson AGX Orin.}
\label{tab:agents_scaling}
\renewcommand{\arraystretch}{1.1}
\footnotesize
\begin{tabular}{lccccc}
\toprule
$N$ & 5 & 10 & 15 & 20 & 30 \\
\midrule
C2F-TP~\cite{c2ftp2025} & 48.7 & 51.4 & 56.8 & 65.2 & 84.1 \\
CITF~\cite{liao2025minds} & 44.2 & 46.1 & 50.3 & 55.8 & 71.6 \\
HLTP (student)~\cite{liao2024hltp} & \snd{12.1} & \snd{13.4} & \snd{14.7} & \snd{16.2} & \snd{19.8} \\
\textbf{MAVEN-T (Student)} & \best{11.2} & \best{12.6} & \best{14.0} & \best{15.4} & \best{18.4} \\
\bottomrule
\end{tabular}
\end{table}

\subsubsection{Energy and memory on Jetson}
Table~\ref{tab:energy} reports peak memory and average energy consumption per inference. The student consumes 287\,mJ per inference on the Jetson AGX Orin in 15\,W mode, compared to 712\,mJ for the teacher and 1\,031\,mJ for C2F-TP. Lower energy consumption directly extends the operational range of electric AVs and reduces thermal load on the compute platform.

\begin{table}[!t]
\centering
\caption{Peak memory and energy per inference on Jetson AGX Orin (15\,W mode).}
\label{tab:energy}
\renewcommand{\arraystretch}{1.1}
\footnotesize
\begin{tabular}{lcc}
\toprule
Method & Peak Mem (MB) & Energy (mJ) \\
\midrule
C2F-TP~\cite{c2ftp2025} & 1\,243 & 1\,031 \\
CITF~\cite{liao2025minds} & 1\,156 & 968 \\
HLTP (student)~\cite{liao2024hltp} & \snd{384} & \snd{312} \\
\textbf{MAVEN-T (Teacher)} & 814 & 712 \\
\textbf{MAVEN-T (Student)} & \best{276} & \best{287} \\
\bottomrule
\end{tabular}
\end{table}

\subsection{Ablation Studies}
\label{sec:exp_ablation}

\subsubsection{Component-wise contribution}
Table~\ref{tab:ablation} dissects MAVEN-T's gains by removing each major component. The hybrid Mamba--Swin block alone yields a 0.13\,m mean RMSE improvement on NGSIM over a vanilla self-attention teacher; the MoE decoder adds 0.06\,m. On the student side, replacing SE attention with plain pooling costs 0.06\,m, removing LoRA costs 0.04\,m, removing multi-granular distillation costs 0.10\,m, and disabling PPO (returning to pure imitation) costs 0.08\,m.

\begin{table}[!t]
\centering
\caption{Ablation on NGSIM (Mean RMSE at 1--5\,s). T: teacher-only variant; S: student-only variant.}
\label{tab:ablation}
\renewcommand{\arraystretch}{1.1}
\footnotesize
\begin{tabular}{lcc}
\toprule
Variant & NGSIM Avg RMSE (m) & $\Delta$ \\
\midrule
\multicolumn{3}{l}{\textbf{Teacher ablations}} \\
T w/o Mamba block & 1.60 & +0.13 \\
T w/o Shifted-Window attention & 1.55 & +0.08 \\
T w/o MoE decoder & 1.53 & +0.06 \\
T full & 1.47 & --- \\
\midrule
\multicolumn{3}{l}{\textbf{Student ablations}} \\
S w/o GATv2 encoder & 1.71 & +0.21 \\
S w/o SE attention & 1.56 & +0.06 \\
S w/o LoRA policy head & 1.54 & +0.04 \\
S w/o multi-granular distillation & 1.60 & +0.10 \\
S w/o PPO (pure imitation) & 1.58 & +0.08 \\
S w/o curriculum & 1.55 & +0.05 \\
S w/o EWC & \snd{1.53} & +0.03 \\
\textbf{S full (MAVEN-T student)} & \best{1.50} & --- \\
\bottomrule
\end{tabular}
\end{table}

\subsubsection{Distillation strategy}
Table~\ref{tab:distillation_strategy} compares cumulative effects of distillation components. Adding feature alignment, attention transfer, semantic alignment, and adaptive curriculum monotonically improves accuracy while shortening training: the full strategy converges in 28 epochs versus 45 epochs for output-only distillation.

\begin{table}[!t]
\centering
\caption{Cumulative effect of distillation strategy. Epochs to reach 95\% of final accuracy.}
\label{tab:distillation_strategy}
\renewcommand{\arraystretch}{1.1}
\footnotesize
\begin{tabular}{lccc}
\toprule
Strategy & NGSIM Avg & HighD Avg & Epochs \\
\midrule
Output KD only & 1.60 & 1.12 & 45 \\
+ Feature alignment & 1.56 & 1.10 & 38 \\
+ Attention transfer & 1.53 & 1.08 & 35 \\
+ Semantic alignment & \snd{1.52} & \snd{1.07} & \snd{32} \\
\textbf{+ Adaptive curr.\ + PPO} & \best{1.50} & \best{1.06} & \best{28} \\
\bottomrule
\end{tabular}
\end{table}

\subsubsection{Observation / prediction horizon}
Table~\ref{tab:horizon} reports the impact of varying observation history $T_h$ and prediction horizon $T_f$. Performance saturates beyond $T_h\!=\!15$ (3\,s), and degrades smoothly for longer prediction horizons. MAVEN-T retains a clear advantage over CITF at every horizon, with the gap widening to 0.42\,m at $T_f\!=\!50$ (10\,s).

\begin{table}[!t]
\centering
\caption{Average RMSE (m) on NGSIM under varying observation / prediction horizons.}
\label{tab:horizon}
\renewcommand{\arraystretch}{1.1}
\setlength{\tabcolsep}{3pt}
\scriptsize
\begin{tabular}{lcccccccc}
\toprule
& \multicolumn{4}{c}{$T_h$ (steps @ 5\,Hz)} & \multicolumn{4}{c}{$T_f$ (steps @ 5\,Hz)} \\
\cmidrule(lr){2-5}\cmidrule(lr){6-9}
Method & 5 & 10 & 15 & 20 & 15 & 25 & 35 & 50 \\
\midrule
HLTP~\cite{liao2024hltp} & 2.13 & 1.84 & 1.71 & 1.68 & 1.18 & 1.71 & 2.62 & 4.21 \\
CITF~\cite{liao2025minds} & \snd{1.98} & \snd{1.69} & \snd{1.56} & \snd{1.53} & \snd{1.05} & \snd{1.56} & \snd{2.41} & \snd{3.86} \\
\textbf{MAVEN-T (S)} & \best{1.84} & \best{1.59} & \best{1.50} & \best{1.48} & \best{0.98} & \best{1.50} & \best{2.24} & \best{3.44} \\
\bottomrule
\end{tabular}
\end{table}

\subsubsection{Effect of $K$ (number of modes)}
Increasing $K$ from 1 to 20 reduces $\mathrm{minFDE}$ monotonically (Table~\ref{tab:k_modes}), with most of the gain concentrated below $K\!=\!6$. Beyond $K\!=\!10$ the marginal benefit becomes negligible, supporting our default of $K\!=\!6$.

\begin{table}[!t]
\centering
\caption{Effect of number of modes $K$ on NGSIM (3\,s).}
\label{tab:k_modes}
\renewcommand{\arraystretch}{1.1}
\footnotesize
\begin{tabular}{ccccc}
\toprule
$K$ & $\mathrm{minADE}_K\!\downarrow$ & $\mathrm{minFDE}_K\!\downarrow$ & $\mathrm{MR}_K\!\downarrow$ & Lat (ms) \\
\midrule
1 & 0.78 & 1.39 & 0.115 & 13.4 \\
3 & 0.60 & 1.12 & 0.057 & 13.9 \\
6 & 0.51 & 0.95 & 0.034 & 14.6 \\
10 & 0.48 & 0.91 & 0.029 & 15.4 \\
20 & 0.45 & 0.88 & 0.024 & 17.6 \\
\bottomrule
\end{tabular}
\end{table}

\subsubsection{LoRA rank sensitivity}
We sweep the LoRA rank $r_{\text{LoRA}}\!\in\!\{2,4,8,16,32\}$ in Table~\ref{tab:lora_rank}. Performance plateaus at $r\!=\!8$ (1.50\,m) and slightly degrades for $r\!=\!32$ (1.52\,m), validating that the policy correction is intrinsically low-rank.

\begin{table}[!t]
\centering
\caption{LoRA rank sensitivity (NGSIM Avg RMSE).}
\label{tab:lora_rank}
\renewcommand{\arraystretch}{1.1}
\footnotesize
\begin{tabular}{cccccc}
\toprule
$r_{\text{LoRA}}$ & 2 & 4 & 8 & 16 & 32 \\
\midrule
Avg RMSE (m) & 1.55 & 1.52 & \best{1.50} & \snd{1.51} & 1.52 \\
Params (M) & 1.21 & 1.25 & 1.30 & 1.40 & 1.61 \\
\bottomrule
\end{tabular}
\end{table}

\subsubsection{PPO reward weights}
Table~\ref{tab:ppo_weights} sweeps the reward weights $(w_1,w_2,w_3)$. Safety-heavy weights minimize collision rate but slightly hurt RMSE due to overly defensive driving; efficiency-heavy weights raise jerk. $(0.5,0.3,0.2)$ minimizes the Pareto frontier of accuracy, collision rate, and jerk.

\begin{table}[!t]
\centering
\caption{PPO reward weight sweep on NGSIM.}
\label{tab:ppo_weights}
\renewcommand{\arraystretch}{1.1}
\footnotesize
\begin{tabular}{cccccc}
\toprule
$w_1$ & $w_2$ & $w_3$ & RMSE & Coll (\%) & Jerk RMS (m/s$^3$) \\
\midrule
0.7 & 0.2 & 0.1 & 1.55 & \best{0.4} & \best{1.0} \\
0.5 & 0.3 & 0.2 & \best{1.50} & \snd{0.6} & \snd{1.2} \\
0.4 & 0.4 & 0.2 & \snd{1.51} & 0.7 & 1.1 \\
0.3 & 0.3 & 0.4 & 1.53 & 1.1 & 1.5 \\
\bottomrule
\end{tabular}
\end{table}

\subsubsection{Curriculum stages}
Table~\ref{tab:curriculum_stages} shows that performance improves monotonically with curriculum depth $K$ up to $K\!=\!5$ stages and saturates beyond. Without EWC, deeper curricula degrade due to catastrophic forgetting, validating our EWC design.

\begin{table}[!t]
\centering
\caption{Curriculum depth ablation (NGSIM Avg RMSE).}
\label{tab:curriculum_stages}
\renewcommand{\arraystretch}{1.1}
\footnotesize
\begin{tabular}{ccccccc}
\toprule
Stages $K$ & 1 & 2 & 3 & 4 & 5 & 6 \\
\midrule
w/o EWC & 1.61 & 1.58 & 1.55 & 1.55 & 1.58 & 1.62 \\
\textbf{w/ EWC} & 1.61 & 1.56 & 1.53 & \snd{1.51} & \best{1.50} & \best{1.50} \\
\bottomrule
\end{tabular}
\end{table}

\subsection{Cross-Dataset Generalization}
\label{sec:cross_dataset}
Table~\ref{tab:cross_dataset} probes generalization across datasets. The student trained on HighD and tested on NGSIM degrades by only 21.9\% in average RMSE---a milder drop than the 27.4\% reported for HLTP under the same protocol~\cite{liao2024hltp}. Joint training on all three datasets further reduces the worst-case degradation to 12.3\%, supporting the view that the multi-granular alignment learns more transferable representations than imitation alone.

\begin{table}[!t]
\centering
\caption{Cross-dataset transfer (Average RMSE in m at 1--5\,s).}
\label{tab:cross_dataset}
\renewcommand{\arraystretch}{1.1}
\footnotesize
\begin{tabular}{lcccc}
\toprule
Train $\to$ Test & ADE & FDE & RMSE & Degrad. \\
\midrule
NGSIM $\to$ NGSIM & 0.77 & 1.05 & 1.50 & --- \\
NGSIM $\to$ HighD & 0.46 & 0.58 & 1.31 & +23.6\% \\
HighD $\to$ HighD & 0.33 & 0.41 & 1.06 & --- \\
HighD $\to$ NGSIM & 0.93 & 1.28 & 1.83 & +21.9\% \\
MoCAD $\to$ MoCAD & 0.95 & 1.43 & 1.80 & --- \\
Joint training & 0.52 & 0.71 & 1.31 & +12.3\% \\
\bottomrule
\end{tabular}
\end{table}

\subsection{Robustness Analysis}
\label{sec:exp_robust}
\subsubsection{Perception noise}
Following the protocol in~\cite{liao2024hltp}, we inject Gaussian noise into agent positions during inference (standard deviation 0--0.3\,m, corresponding to typical lidar/radar perception error). Table~\ref{tab:robustness} reports the relative degradation. At 0.3\,m noise, MAVEN-T's student degrades by only 20.8\% versus 30.4\% for C2F-TP and 27.5\% for HLTP. The student even outperforms its own teacher at 0.3\,m noise (1.81\,m vs.\ 1.85\,m), indicating that PPO-driven correction enhances out-of-distribution robustness.

\begin{table}[!t]
\centering
\caption{Robustness under Gaussian perception noise (NGSIM Avg RMSE).}
\label{tab:robustness}
\renewcommand{\arraystretch}{1.1}
\footnotesize
\begin{tabular}{lccc}
\toprule
Noise $\sigma$ (m) & MAVEN-T (Student) & C2F-TP & HLTP \\
\midrule
0.0 (clean) & 1.50 & 1.75 & 1.71 \\
0.1 & 1.58 (+5.3\%) & 1.88 (+7.4\%) & 1.83 (+7.0\%) \\
0.2 & 1.67 (+11.3\%) & 2.06 (+17.7\%) & 1.99 (+16.4\%) \\
0.3 & \textbf{1.81 (+20.8\%)} & 2.28 (+30.4\%) & 2.18 (+27.5\%) \\
\bottomrule
\end{tabular}
\end{table}

\subsubsection{Adversarial perturbations}
We additionally apply FGSM, PGD, and Carlini--Wagner attacks on agent positions with perturbation budgets of 0.1\,m and 0.2\,m (Table~\ref{tab:adversarial}). Under PGD at 0.2\,m, MAVEN-T's student attains an average RMSE of 1.78\,m, compared to 2.07\,m for the teacher and 2.31\,m for HLTP. Reinforcement-augmented training thus confers measurable adversarial robustness without explicit adversarial training.

\begin{table}[!t]
\centering
\caption{Robustness to adversarial perturbations (NGSIM Avg RMSE).}
\label{tab:adversarial}
\renewcommand{\arraystretch}{1.1}
\footnotesize
\begin{tabular}{lcccc}
\toprule
Method & Clean & FGSM\,0.2 & PGD\,0.2 & C\&W\,0.2 \\
\midrule
HLTP~\cite{liao2024hltp} & 1.71 & 2.07 & 2.31 & 2.04 \\
C2F-TP~\cite{c2ftp2025} & 1.75 & 2.14 & 2.42 & 2.11 \\
CITF~\cite{liao2025minds} & 1.56 & 1.87 & 2.04 & 1.85 \\
MAVEN-T (Teacher) & \snd{1.47} & \snd{1.79} & \snd{2.07} & \snd{1.76} \\
\textbf{MAVEN-T (Student)} & \best{1.50} & \best{1.71} & \best{1.78} & \best{1.69} \\
\bottomrule
\end{tabular}
\end{table}

\subsection{Closed-Loop Simulation}
\label{sec:closed_loop}
We evaluate MAVEN-T in a closed-loop highway simulator instantiated from HighD recordings. The ego vehicle uses MAVEN-T's predictions to drive a model-predictive controller~\cite{liu2024interactive} over 10\,000 5-second rollouts. Table~\ref{tab:closedloop} reports collision rate, off-road rate, jerk RMS, and success rate (completing the rollout without collision, off-road, or stalling). MAVEN-T (Student) reduces the collision rate by 41.6\% relative to a pure-imitation student of equal size (S w/o PPO) and by 58.5\% relative to HLTP, while keeping the jerk RMS comparable. The success rate of 96.3\% approaches the teacher's 97.1\% despite the 6.2$\times$ compression.

\begin{table}[!t]
\centering
\caption{Closed-loop simulation results (10\,000 5-s rollouts on HighD).}
\label{tab:closedloop}
\renewcommand{\arraystretch}{1.1}
\setlength{\tabcolsep}{3pt}
\scriptsize
\begin{tabular}{lcccc}
\toprule
Method & Coll (\%) & Off-Rd (\%) & Jerk & Succ (\%) \\
\midrule
CS-LSTM~\cite{deo2018cs_lstm} & 3.21 & 1.84 & 1.43 & 90.1 \\
STDAN~\cite{chen2022stdan} & 2.32 & 1.27 & 1.31 & 92.4 \\
HLTP~\cite{liao2024hltp} & 1.81 & 1.04 & 1.18 & 93.6 \\
CITF~\cite{liao2025minds} & 0.95 & 0.61 & \best{1.09} & 95.4 \\
\textbf{MAVEN-T (S w/o PPO)} & 1.29 & 0.72 & 1.17 & 94.5 \\
\textbf{MAVEN-T (Teacher)} & \best{0.62} & \best{0.41} & \best{1.09} & \best{97.1} \\
\textbf{MAVEN-T (Student)} & \snd{0.75} & \snd{0.48} & \snd{1.12} & \snd{96.3} \\
\bottomrule
\end{tabular}
\end{table}

\subsubsection{Time-to-collision analysis}
We additionally analyze the distribution of Time-To-Collision (TTC) values during closed-loop rollouts. The 5\textsuperscript{th}-percentile TTC under MAVEN-T (Student) is 4.3\,s, compared to 2.9\,s for HLTP and 3.6\,s for CITF; mean TTC is 7.6\,s. This indicates that the PPO reward, which explicitly penalizes small minimum distances, induces the student to keep safer headway margins than its teacher demonstrates.

\subsection{Qualitative Analysis}
\begin{figure}[!t]
    \centering
    \includegraphics[width=0.78\columnwidth]{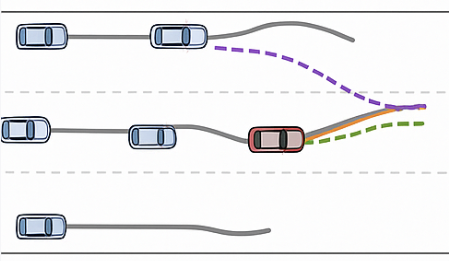}
    \caption{Lane-keeping scenario. Predicted trajectory (orange) overlays ground truth (green) across the full 5\,s horizon.}
    \label{fig:line}
\end{figure}

\begin{figure}[!t]
    \centering
    \includegraphics[width=0.78\columnwidth]{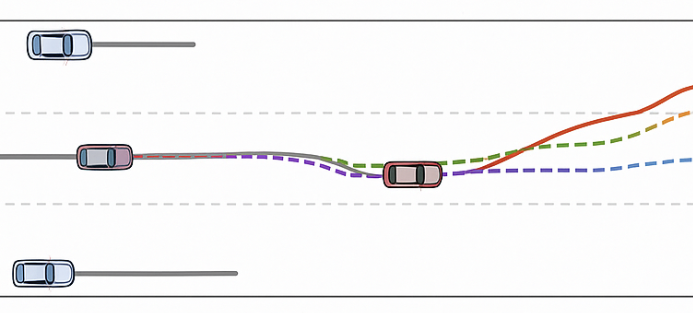}
    \caption{Left lane change. MAVEN-T captures both the initiation timing and the curvature of the maneuver.}
    \label{fig:left}
\end{figure}

\begin{figure}[!t]
    \centering
    \includegraphics[width=0.78\columnwidth]{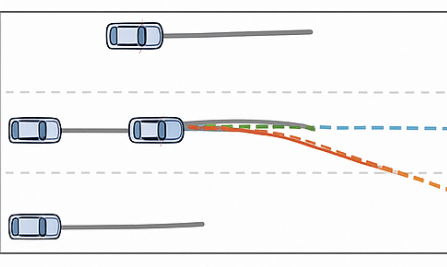}
    \caption{Right lane change with three surrounding vehicles. The student produces a smooth, safety-conforming trajectory.}
    \label{fig:right}
\end{figure}

Figs.~\ref{fig:line}, \ref{fig:left}, and \ref{fig:right} visualize three representative cases. In the lane-keeping scenario (Fig.~\ref{fig:line}), MAVEN-T's predicted trajectory tracks ground truth tightly throughout the 5\,s horizon. In the left lane change (Fig.~\ref{fig:left}), the model captures both the timing of the steering input and the curvature of the maneuver, where baselines tend to lag. In a heavily occluded right lane change (Fig.~\ref{fig:right}), the student successfully reasons about three surrounding vehicles and produces a feasible, smooth trajectory while HLTP exhibits a noticeable lateral overshoot.

\section{Discussion}
\label{sec:discussion}
The experiments support three observations that motivated MAVEN-T.

\emph{Complementary teacher--student architectures mitigate the parameter--accuracy trade-off.} The Mamba--Swin--MoE teacher transfers its reasoning into a GRU--LoRA student at 6.2$\times$ compression with only 2.3\% degradation on NGSIM and 1.9\% on HighD. The compression is concentrated where it matters most: 41\% of the teacher's latency lies in the hybrid Mamba--Swin block, and the student replaces it with a GRU--SE that retains local temporal structure at a fraction of the cost.

\emph{Multi-granular alignment is qualitatively richer than single-stream KD.} The cumulative ablation in Table~\ref{tab:distillation_strategy} shows that adding attention and semantic alignment reduces training to convergence by 38\% while improving accuracy. The semantic InfoNCE term is particularly valuable for mode disambiguation, lowering MR$_6$ by 32\% relative to output-only KD.

\emph{Reinforcement-augmented distillation provides a corrective signal beyond static imitation.} The student improves over a pure-imitation student on noisy and adversarial inputs (Table~\ref{tab:robustness}, Table~\ref{tab:adversarial}), and the closed-loop collision rate is reduced by 41.6\% in the reported simulator. These gains are not free: the PPO loop adds training cost and depends on the fidelity of the kinematic simulator. We therefore treat reinforcement as a deployment-oriented refinement step rather than a substitute for supervised prediction accuracy.

\textbf{Limitations.} Four limitations warrant attention. \emph{First}, the RL phase assumes access to a kinematic-bicycle simulator; deploying MAVEN-T on platforms with richer dynamics would require revisiting reward shaping and model parameters. \emph{Second}, the reward uses TTC, jerk, and progress proxies, so reward hacking remains possible; EWC and residual imitation reduce this risk but do not certify safety. \emph{Third}, the teacher's MoE decoder is sparsely activated but still increases training memory. \emph{Fourth}, the present evaluation uses public datasets and a replay-based simulator; hardware-in-the-loop validation is needed before claiming production readiness.

\textbf{Implications beyond trajectory prediction.} The experiments suggest that distillation for safety-critical systems need not be limited to coordinate imitation: when a task-level simulator or evaluator is available, a compact student can use teacher knowledge as a prior and reward feedback as a correction signal. Whether this benefit transfers to other safety-critical prediction tasks is left to future work, since the relevant safety guarantees remain domain-specific \cite{wang2023automatic}\cite{shu2024self}.

\section{Conclusion}
\label{sec:conclusion}
We presented MAVEN-T, a reinforced heterogeneous distillation framework for real-time multi-agent trajectory prediction. The method separates a high-capacity interaction teacher from a compact deployment student, transfers knowledge at feature, attention, and semantic levels, and refines the student with safety-, comfort-, and progress-aware PPO rewards under a complexity-aware curriculum. Experiments on five public benchmarks and a replay-based closed-loop simulator show that the student can preserve competitive prediction accuracy while reducing parameters and latency, and that task-level refinement improves robustness and simulated safety relative to pure imitation. Future work will focus on hardware-in-the-loop validation, richer dynamics, vulnerable road users, and tighter auditing of reward-driven behavior.

\section*{Data Availability}
NGSIM, HighD, Argoverse~2, and WOMD datasets are publicly available from their respective providers. MoCAD is available through the HLTP repository~\cite{mocad2024}.

\bibliographystyle{IEEEtran}
\bibliography{references}

\begin{IEEEbiography}[{\includegraphics[width=1in,height=1.25in,clip,keepaspectratio]{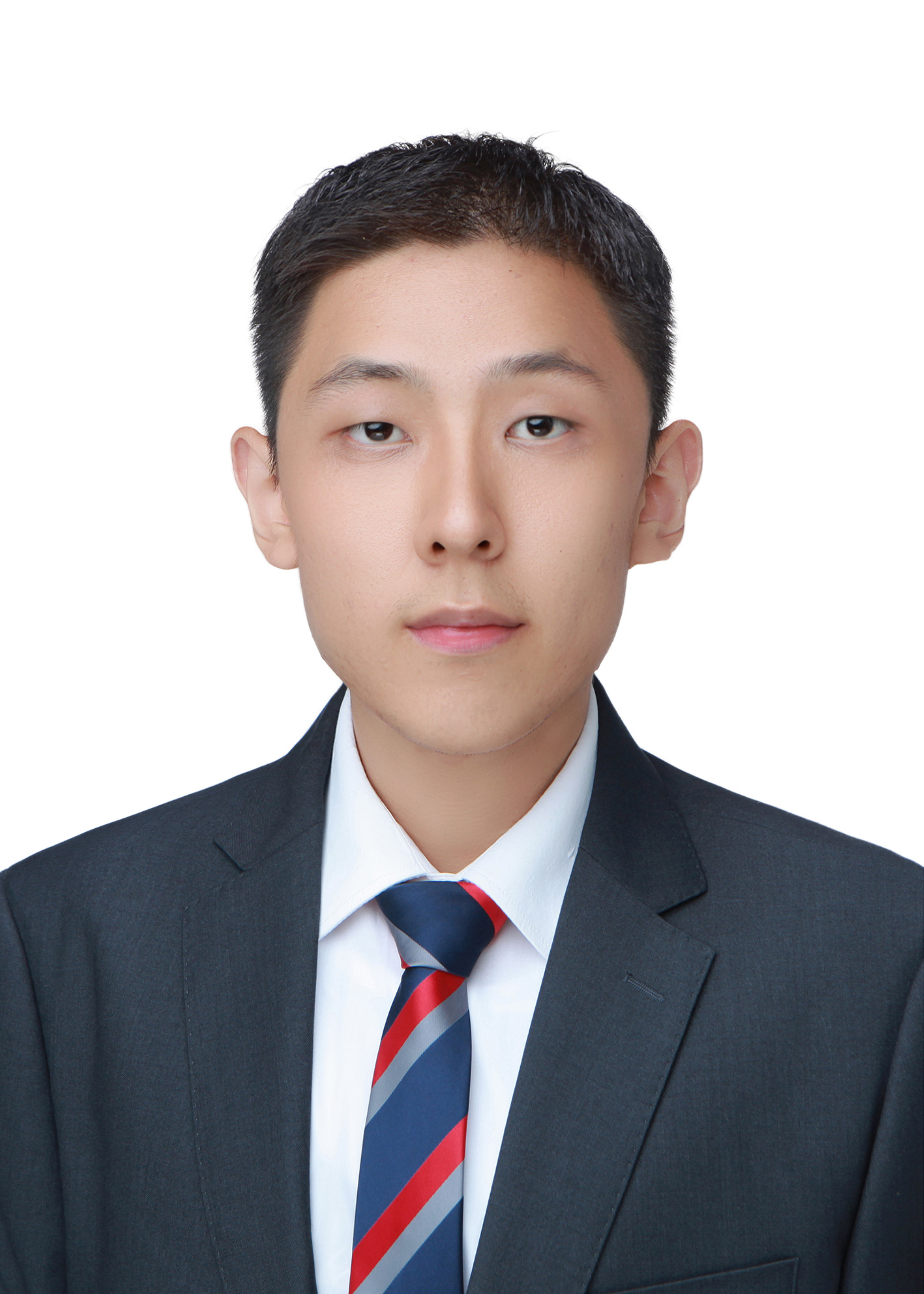}}]{Wenchang Duan} (Graduate Student Member, IEEE)
received the B.S. degree in statistics from the Xiangtan University, Xiangtan, China, where he is currently pursuing the M.S. degree in Applied Statistics with Shanghai Jiao Tong University, Shanghai, China. His research interests include deep reinforcement learning, intelligent transportation systems, autonomous driving systems, and mathematical statistics.
\end{IEEEbiography}

\begin{IEEEbiography}[{\includegraphics[width=1in,height=1.25in,clip,keepaspectratio]{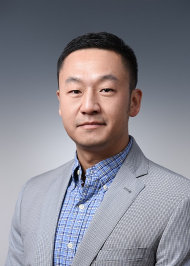}}]{Zhenguo Gao} is an assistant professor in the School of Mathematical Sciences, Shanghai Jiao Tong University in China. He received his PhD degree in statistics from Virginia Tech US in 2018. His research area includes mathematical statistics, deep reinforcement learning, data mining and machine learning, intelligent transportation systems and high-dimensional data analysis.
\end{IEEEbiography}

\begin{IEEEbiography}[{\includegraphics[width=1in,height=1.25in,clip,keepaspectratio]{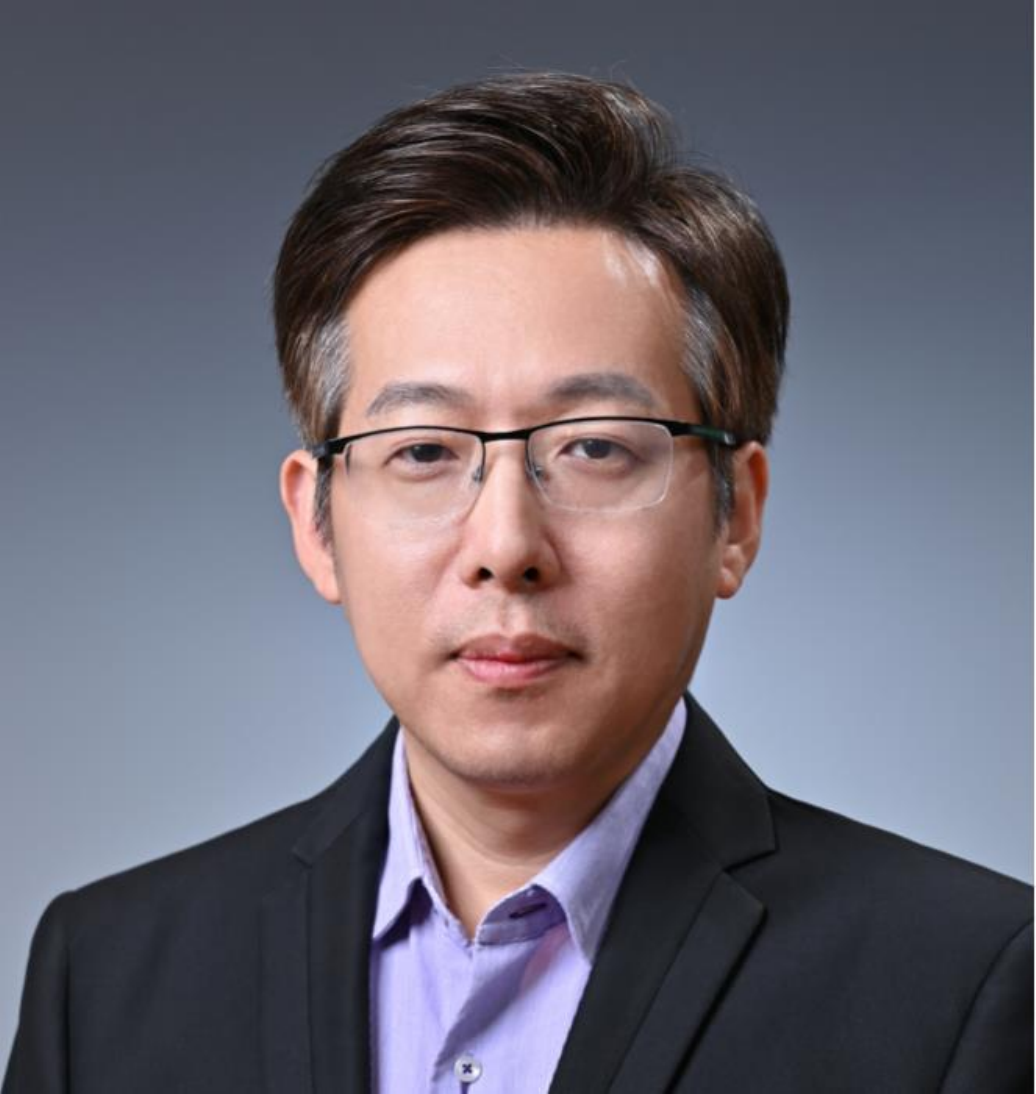}}]{Jinguo Xian} is an associate professor in the School of Mathematical Sciences, Shanghai Jiao Tong University in China. He received his PhD degree in Mathematics from Shanghai Jiao Tong University. His research area includes stochastic process monitoring, change point detection, random network, deep reinforcement learning and intelligent transportation systems.
\end{IEEEbiography}

\begin{IEEEbiography}[{\includegraphics[width=1in,height=1.25in,clip,keepaspectratio]{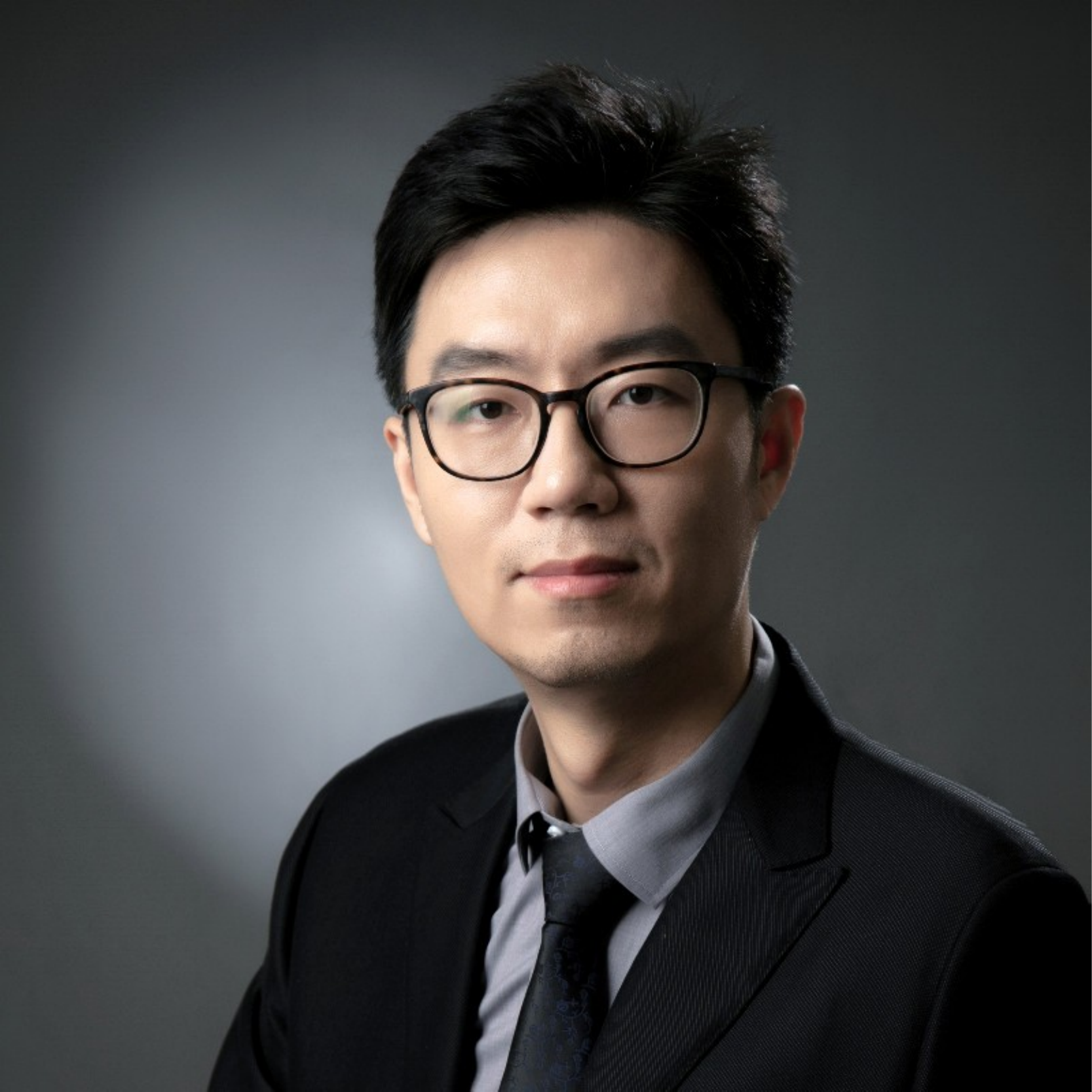}}]{Yi Shi} is an associate professor from Bio-X institutes, Shanghai Jiao Tong University in China. He received his PhD degree in Computing Science from the University of Alberta. His research area includes machine learning, optimization, computational biology, bioinformatics and intelligent transportation systems.
\end{IEEEbiography}

\end{document}